\documentclass[journal]{IEEEtran}
\ifCLASSINFOpdf
\else
\fi
\usepackage{epsfig}
\usepackage{epstopdf}
\usepackage{graphicx}
\usepackage{enumerate}
\usepackage{amsmath}
\usepackage{amssymb}
\usepackage[ruled,vlined]{algorithm2e}
\usepackage{bbm}
\usepackage{listings}% http://ctan.org/pkg/listings
\usepackage{subcaption}
\usepackage{kantlipsum} %<- For dummy text
\usepackage{mwe} %<- For dummy images

\newtheorem{theorem}{Theorem}
\newtheorem{remark}[theorem]{Remark}
\newtheorem{corollary}[theorem]{Corollary}

\usepackage[ruled,vlined]{algorithm2e}
\usepackage{hyperref}

\usepackage{color}

\begin{document}
%\title{Weighted Low-Rank Approximation for Background Modelling}
\title{Weighted Low-Rank Approximation\\ of Matrices and Background Modeling}

\author{Aritra Dutta, %~\IEEEmembership{Member,~IEEE,}
        Xin Li, %~\IEEEmembership{Fellow,~OSA,}
        and~Peter Richt\'{a}rik%~\IEEEmembership{Life~Fellow,~IEEE}% <-this % stops a space
\thanks{Aritra Dutta is with the Visual Computing Center, Division of Computer, Electrical and Mathematical Sciences and Engineering (CEMSE) at King Abdullah University of Science and Technology, Thuwal, Saudi Arabia-23955-6900, 
e-mail: aritra.dutta@kaust.edu.sa (see https://aritradutta.weebly.com/)}% <-this % stops a space
\thanks{Xin Li is with the Department of Mathematics, University of Central Florida, FL, USA-32816, email: xin.li@ucf.edu.}
\thanks{Peter Richt\'{a}rik is with the Visual Computing Center, Division of Computer, Electrical and Mathematical Sciences and Engineering (CEMSE) at King Abdullah University of Science and Technology, University of Edinburgh, and MIPT, 
e-mail: peter.richtarik@kaust.edu.sa (see http://www.maths.ed.ac.uk/~prichtar/).}% <-this % stops a space
}

\maketitle

\begin{abstract}
We primarily study a special a weighted low-rank approximation of matrices and then apply it to solve the background modeling problem. We propose two algorithms for this purpose: one operates in the batch mode on the entire data and the other one operates in the batch-incremental mode on the data and naturally captures more background variations and computationally more effective. Moreover, we propose a robust technique that learns the background frame indices from the data and does not require any training frames. We demonstrate through extensive experiments that by inserting a simple weight in the Frobenius norm, it can be made robust to the outliers similar to the $\ell_1$ norm. Our methods match or outperform several state-of-the-art online and batch background modeling methods in virtually all quantitative and qualitative measures.
\end{abstract}


% \fi
\IEEEpeerreviewmaketitle

\vspace{-0.1in}

\section{Introduction}

\IEEEPARstart{W}e give a brief review of the classical low rank approximation of matrices and introduce the background modeling problem. 

\vspace{-0.15in}
\subsection{Low-rank approximation} %Low-rank approximation (PCA) as a linear method.

The standard low rank approximation aka the principal component analysis (PCA) problem can be defined as an approximation to a given matrix $A\in\mathbb{R}^{m\times n}$ by a rank $r$ matrix under the Frobenius norm:\begin{eqnarray}
\label{pca}
X^*=\arg\min_{\substack{{X}\in\mathbb{R}^{m\times n}\\{\rm rank}({X})\le r}}\|A-{X}\|_F^2,
\end{eqnarray}
where $\|\cdot\|_F$ denotes the Frobenius norm of matrices. The solutions to~(\ref{pca}) are given by
\begin{align}\label{hardthresholding}
X^*=H_r(A):=U\Sigma_rV^T,
\end{align}
where $A$ has singular value decompositions 
$
A=U\Sigma V^T,
$
and $\Sigma_r(A)$ is the diagonal matrix obtained from $\Sigma$ by hard-thresholding operation that keeps only $r$  largest singular values and replaces the other singular values by 0 along the diagonal. This is also referred to as Eckart-Young-Mirsky's theorem~(\cite{golub}) and is closely related to the PCA method in statistics~\cite{pca}. In image processing, rank-reduced signal processing, computer vision, and in many other engineering applications SVD is a successful dimension reduction tool. 
 The low rank matrix obtained through PCA is a good approximation to the data matrix $A$ if $A$ contains only normally (and independently) distributed noise. But, in many real world problems, if sparse large errors or outliers are present in the data matrix, PCA fails to deal with it and thus additional regularization has been introduced to accommodate the sparse outliers.  One can think the background modeling problem in video sequences as a good example of such a real world problem. In the next section we will review the classic background modeling problem in light of matrix decomposition and briefly survey several state-of-the-art algorithms used to solve it. 

\vspace{-0.15in}
\subsection{Background modeling in matrix decomposition framework}\label{others}
Background modeling and moving object detection are two key steps in many computer vision systems and video-surveillance applications. In the past decade, one of the most prevalent approaches used in background estimation is to treat it as a matrix decomposition problem (\cite{Bouwmans201431,Bouwmans2016,Sobral20144}). Given a sequence of $n$ video frames with each frame converted into a vector ${\mathbf a}_i\in {\mathbb R}^m$, $i=1,2,...,n$,~the data matrix $A=({\mathbf a}_1, {\mathbf a}_2, ... , {\mathbf a}_n)\in {\mathbb R}^{m\times n}$ is the collection of all the frame vectors. Therefore, it is natural to consider a matrix decomposition problem by decomposing $A$ as the sum of its background and foreground:
\begin{eqnarray*}
A=B+F,
\end{eqnarray*}
~\\[-0.15in]
where $B,F\in {\mathbb R}^{m\times n}$ are the background and foreground matrices, respectively. The above problem is ill-posed, and it requires more information about the structure of the decomposition. In practice, the background $B$ is expected to stay static or close to static throughout the frames when the camera motion is small and so $B$ is assumed to be low-rank~\cite{oliver1999}. At the same time, the foreground, $F$ is usually sparse if its size is relatively small compared to the frame size (\cite{Bouwmans201431,Bouwmans2016,candeslimawright,APG,LinChenMa}). These and similar observations leads to the models of the form (\cite{Bouwmans201431,Bouwmans2016,APG,LinChenMa,duttaligongshah,RPCA-BL,prmf,grasta,gosus,dutta_thesis}):
\begin{equation}\label{eq:general}\min_{\substack{B, F\\ A=B+F}} f_{\rm rank}(B) + f_{\rm sparse}(F),\end{equation}
~\\[-0.12in]
where $f_{\rm rank}$ is a function that encourages the rank of $B$ to be low, and $f_{\rm sparse}$ is a function that encourages the foreground $F$ to be sparse. 
Next we will discuss how this idea transformed into several state-of-the-art algorithms to solve the background modeling problem.  

\paragraph{Robust principal component analysis (RPCA)} By using the above idea, RPCA~\cite{candeslimawright,LinChenMa,APG} was introduced to solve the background modeling problem by considering the background frames, $B$, having a low-rank structure and the foreground,~$A-B$,~sparse. The nuclear norm~(sum of the singular values) is used on background matrix $B$ as a surrogate of rank and the $\ell_1$ norm is used to encourage sparsity in the foreground:
\begin{equation}\label{rpca}
\min_B\|A-B\|_{\ell_1}+\lambda \|B\|_*,
\end{equation}
where $\|\cdot\|_{\ell_1}$ and $\|\cdot\|_*$ denote the $\ell_1$ norm and the nuclear norm of matrices, respectively. RPCA is considered to be one of the most widely used state-of-art approaches to solve the background modeling problem. Several algorithms have been proposed to solve RPCA \cite{candeslimawright,LinChenMa,APG}, for example, inexact and exact augmented Lagrange method (iEALM and EALM), accelerated proximal gradient (APG), just to name a few. However, RPCA cannot take advantage of any possible extra prior information on the background and can not be used as a supervised learning method. 

\paragraph{Generalized fused Lasso~(GFL)} Consider a supervised learning situation when some pure background frames are given. Recently in \cite{xin2015}, Xin et. al. proposed a supervised model called the GFL to address these issues. Let the data matrix $A$ be written into $A=(A_1 ~A_2)$, where $A_1$ contains the given pure background frames. Xin et. al. ~\cite{xin2015} proposed the following model: Find the background matrix $B=(B_1~B_2)$, and foreground matrix $F=(F_1 ~F_2)$, partitioned in the same way as $A$, such that
\begin{align}
	\min_{\substack{B,F\\B_1=A_1\\A_2=B_2+F_2}}{\rm rank}(B)+\|F\|_{GFL},
\end{align}
where $\|\cdot\|_{GFL}$ denotes a norm that combines the $\ell_1$ norm and a local spatial total variation norm (to encourage connectivity of the foreground). To make the problem more tractable, Xin et. al. further specialized the above model by assuming ${\rm rank}(B)={\rm rank}(B_1)$. Since $B_1=A_1$ and $A_1$ is given, so $r:={\rm rank}(B_1)$ is also given and thus, we can re-write the model of \cite{xin2015} as a special case of the following:
\begin{equation}\label{gfl}
	\min_{\substack{B = (B_1\;B_2)\\{\rm rank}(B)\le r\\B_1=A_1}} \|A-B\|_{GFL}.
\end{equation}
It is obvious that, except in different norms, problem \eqref{gfl} is a special constrained (weighted) low-rank approximation problem as in (\ref{golub's problem}).

\paragraph{Incremental methods} 
Conventional PCA~\cite{pca} is an essential tool to numerically solve both RPCA and GFL problems.~PCA operates at a cost of $\min\{\mathcal{O}(m^2n),\mathcal{O}(mn^2)\}$, which is due to the SVD of an $m\times n$ data matrix.~For RPCA algorithms, the space complexity of an SVD computation is approximately~$\mathcal{O}((m+n)r)$, where $r$ is the rank of the low-rank approximation matrix in each iteration, which is increasing.~And a high resolution video sequence characterized by very large $m$, is computationally extremely expensive for the RPCA and GFL algorithms.~For example, APG algorithm runs out of memory to process 600 video frames each of size $300\times 400$ on a computer with 3.1 GHz Intel Core i7-4770S processor and 8GB memory.~In the past few decades, incremental PCA~(IPCA) was developed for machine learning applications to reduce the computational complexity of performing PCA on a huge data set.~The idea is to produce an efficient SVD calculation of an augmented matrix of the form $[A\;\tilde{A}]$ by using the SVD of $A$, where $A\in\mathbb{R}^{m\times n}$ is the original matrix and $\tilde{A}$ contains $r$ newly added columns~\cite{incpca}. Similar to the IPCA, several methods have been proposed to solve the background estimation problem in an incremental manner~\cite{mmb,inrpca}. 

\paragraph{Grassmannian robust adaptive subspace estimation (GRASTA)} In 2012, He et. al. ~\cite{grasta} proposed GRASTA, a robust subspace tracking algorithm and showed its application in background modeling problems. At each time step $i$, GRASTA solves the following optimization problem:~For a given orthonormal basis $U_{\Omega_s}\in\mathbb{R}^{|{\Omega_s}|\times d}$ solve
\begin{align}\label{Grasta}
\min_x\|U_{\Omega_s}x-a_i^{\Omega_s}\|_{\ell_1},
\end{align}
where each video frame $a_i\in\mathbb{R}^m$ is subsampled over the index set ${\Omega_s}\subset\{1,2,\cdots,m\}$ following the model: $a_i^{\Omega_s}=U_{\Omega_s}x+f_i^{\Omega_s}+\epsilon_{\Omega_s}$,~such that, $x\in\mathbb{R}^{d}$ is a weight vector and $\epsilon_{\Omega_s}\in\mathbb{R}^{|{\Omega_s}|}$ is a Gaussian noise vector. After updating $x$, one has to update $U_{\Omega_s}$. 
%{\color{red} What is $b$?}

\paragraph{Recursive projected compressive sensing~(ReProCS)} In 2014,~Guo et. al.~\cite{reprocs} proposed another online algorithm for separating sparse and low dimensional subspace~(see also~\cite{pracreprocs,modified_cs}). ReProCS is a two stage algorithm. In the first stage, given a sequence of training background frames, say $t$, the algorithm finds an approximate basis which is ideally of low-rank.~After estimating the initial low-rank subspace in the second stage, the algorithm recursively estimates $F_{t+1}, B_{t+1}$ and the subspace in which $B_{t+1}$ lies.

\paragraph{Incremental principal component pursuit~(incPCP)}
Rodriguez et. al. ~\cite{incpcp} formulated the incPCP algorithm which processes one frame at a time incrementally and uses only a few frames for initialization of the prior~(see also~\cite{matlab_pcp,inpcp_jitter}). incPCP follows a modified framework of conventional PCP but is built with the assumption that the partial rank $r$ SVD of first $k-1$ background frames $B_{k-1}$ is known. And by using them, $A_{k-1}$ can be written as $A_{k-1}=B_{k-1}+F_{k-1}$.~Therefore, for a new video frame $a_k$, one can solve the optimization problem as follows:
\begin{align}\label{reprocs}
&\min_{\substack{B_k,F_k\\{\rm rank}(B_k)\le r}}\|B_k+F_k-A_k\|_F^2+\lambda\|F_k\|_{\ell_1},
\end{align}
where $A_k = [A_{k-1}~~a_k]$ and $B_k = [U_r\Sigma_rV_r^T~~b_k]$ such that $U_r\Sigma_rV_r^T$ is a partial SVD of $B_{k-1}.$ According to~\cite{incpcp}, the initialization step can be performed incrementally. 

\paragraph{Detecting contiguous outliers in the low-rank representation (DECOLOR)} In DECOLOR, Zhou et. al. \cite{decolor} combined three models: a low-rank background model, energy of the foreground support, $S$, generated from an Ising model, and a signal model that describes the data $A$, given $B$ and $F$. They proposed to minimize the function:
\begin{eqnarray}\label{decolor_problem}
\min_{\substack{B,W\\{\rm rank}(B)\le r\\W_{ij}\in\{0,1\}}}\|(\mathbbm{1}-W)\odot (A-B)\|_F^2+\beta\|W\|_{\ell_1}\nonumber\\+\gamma\|I_{S,\mathcal{V}}{\rm vec}(W)\|_{\ell_1},
\end{eqnarray}
where $W$ is an indicator matrix whose $(i,j)$th entry is 1 if $(i,j)$ represents a foreground pixel and 0 if it corresponds to a background pixel, $\mathbbm{1}$ is the matrix of all 1s, and $I_{S,\mathcal{V}}$ is the incidence matrix of the graph $\mathcal{V}$ that denotes all pixels in the foreground.

\paragraph{Probabilistic approach to robust matrix factorization (PRMF)} In PRMF, Wang et. al. \cite{prmf} decomposed $A$ as: $A=UV^T+F$, where $U$ and $V$ are two rank $r$ matrices. They solved the $\ell_1$ loss function coupled with two regularizer terms:
\begin{eqnarray}\label{prmf_problem}
\min_{U,V}\|W\odot (A-UV^T)\|_{\ell_1}+\frac{\lambda_U}{2}\|U\|_F^2+\frac{\lambda_V}{2}\|V\|_F^2,
\end{eqnarray}
by using the Bayesian perspective and an expectation-maximization approach. Wang et. al. also assumed that each element of the foreground $F$ is sampled independently from the Laplace distribution. In their model, the matrix $W$ is an indicator matrix similar to DECOLOR and the $\ell_2$ regularization terms prevent overfitting. 

\vspace{-0.1in}
\section{Weighted Low-rank Approximation of Matrices}% and background estimation}
The algorithms reviewed in the previous section tried to improve the model given by \eqref{pca} via deviating from the Frobenius norm in order to encourage the detection of outliers in the data matrix. Note that the solutions to \eqref{pca} as given in~(\ref{hardthresholding}) may also suffer from another shortcoming: the fact that none of the entries of $A$ is guaranteed to be preserved in $X^*$. %In many applications this could be a typical weak point of  PCA.
 Indeed, in many real world problems, one has good reasons to keep certain entries of $A$ unchanged while looking for a low rank approximation. In 1987, Golub, Hoffman, and Stewart required that certain columns, $A_1,$ of $A$ must be preserved when one looks for a low rank approximation of $(A_1\;\;A_2)$ and considered the following {\it constrained} low rank approximation problem~\cite{golub}: Given $A=(A_1\;\;A_2)\in\mathbb{R}^{m\times n}$ with $A_1\in\mathbb{R}^{m\times k}$ and $A_2\in\mathbb{R}^{m\times (n-k)}$, find $\tilde{A}_2$ such that~(with $\tilde{A}_1=A_1$)
\begin{eqnarray}\label{golub's problem}
(\tilde{A}_1\;\tilde{A}_2)=\arg\min_{\substack{X_1,X_2\\{\rm rank}(X_1\;\;X_2)\le r\\X_1=A_1}}\|(A_1\;\;A_2)-(X_1\;\;{X}_2)\|_F^2.
\end{eqnarray}

Inspired by applications in which $A_1$ may contain noise, it makes more sense if we require $\|A_1-X_1\|_F$ small instead of looking for $X_1=A_1$. This leads us to consider the following problem: Let $\eta>0$, find $(\hat{X}_1\;\;\hat{X}_2)$ such that
\begin{eqnarray}\label{closeness problem}
(\hat{X}_1\;\;\hat{X}_2)=\displaystyle{\arg\min_{\substack{X_1,X_2\\ \|A_1-X_1\|_F\leq \eta\\{\rm rank}(X_1\;\;X_2)\le r}}\| (A_1\;\;A_2)-({X}_1\;\;{X}_2) \|_F^2.}
\end{eqnarray}
Or, for a large parameter $\lambda $, consider
\begin{eqnarray*}\label{unconstraint closeness}
(\hat{X}_1\;\;\hat{X}_2)=\arg\min_{\substack{X_1,X_2\\{\rm rank}(X_1\;\;X_2)\le r}}\lambda^2\|A_1-X_1\|_F^2+\| A_2-{X}_2 \|_F^2.
\end{eqnarray*}
We observed that the above problem is contained in the following more general point-wise weighted low rank approximation problem:
\begin{eqnarray}\label{hadamard problem}
\boxed{\min_{\substack{X_1,X_2\\{\rm rank}(X_1\;\;X_2)\le r}}\|\left((A_1\;\;A_2)-({X}_1\;\;{X}_2)\right)\odot(W_1\;~W_2)\|_F^2,}
\end{eqnarray}
for $W_1=\lambda \mathbbm{1}$ and $W_2=\mathbbm{1}$ (a matrix whose entries are equal to 1), where $W\in\mathbb{R}^{m\times n}$ is a weight matrix and $\odot$ denotes the Hadamard product.
This block structure in weight matrix, where very few entries are heavily weighted and most entries stay at 1 (unweighted), is realistic in many applications. For example, in the problem of background modeling from a video sequence, each frame is a column in the data matrix and the background is a low rank (ideally of rank 1) component of the data matrix. Therefore, the weight is used to single out the columns that are more likely to be the basis of background frames and the low rank constraint enforces the search for other frames that are in the background subspace. Recent investigations in~\cite{duttaligongshah, duttali_acl} have shown that the above ``approximately preserving" (controlled by a parameter $\lambda$) weighted low rank approximation can be more effective in solving the background modeling, shadows and specularities removal, and domain adaptation problems in computer vision and machine learning. Dutta et. al. in \cite{duttali_acl} showed that if one has prior information about certain frames of the input video matrix as pure background, then one may insist on preserving those columns when looking for a low rank approximation. They reformulated the problem \eqref{hadamard problem} with the weight $W_1$ in the first block to be 
chosen entry-wise rather than scaled by a single constant $\lambda$. This gives more freedom to pixel-wise control of the columns of the given matrix to be preserved. 
%\vspace{-0.12in}
%\subsection{Contributions}
In this paper, (i) we provide a consolidated treatment of our recent contributions to the background modeling problem previously proposed in several conference proceeding papers based on the method that can be categorized as weighted low rank approximation of matrices; and (ii) we present more systematical experiments and a thorough comparison of our proposed algorithms against several state-of-the-art background modeling methods on the mainstream data-sets including Stuttgart, I2R, Wallflower, CDNet 2014, and the SBI data-set,\cite{cvpr11brutzer,lidata,wallflower,cdnet,sbi_a,SBI_web}.

Here is the more detailed outline of the rest of the paper. 

We first discuss an algorithm (Algorithm \ref{wlr}) to solve \eqref{hadamard problem} numerically when  $(W_1)_{ij}\in [\alpha,\beta]$ and $W_2=\mathbbm{1}$. This serves two purposes: that i) our algorithm takes advantage of the block structure of the weights that results in efficient numerical performance as compared to the existing algorithms for solving the general weighted low-rank approximation problem \cite{duttali}, and that ii) we have detailed convergence analysis for the algorithm. 

Next, we review a batch background estimation model in Algorithm \ref{bg_wlr} that is built on Algorithm \ref{wlr}. More specifically, unlike the models described previously (such as in GFL and ReProCS methods reviewed in section~\ref{others}), we do not require any training frames but our algorithm can learn the background frame indices robustly. 
%It uses the entire video sequence to coarsely estimate the frame indices that is a contender of the background. 

Finally, we investigate an adaptive batch-incremental model (Algorithm \ref{inwlr}). Our model finds the background frame indices robustly and incrementally to process the entire video sequence adaptively by going through a sequence of small batches. Therefore, unlike Algorithm \ref{bg_wlr}, Algorithm \ref{inwlr} does not use the entire sequence at once. Rather it operates on local small batch of video frames and use the background information in nearby frames. As a result, it is more time and memory efficient.  %We note that in both models we use the WLR algorithm of Dutta et al. to gain robustness. 

We note that similar to the $\ell_1$ norm used in conventional and in the incremental methods, a weighted Frobenius norm used in Algorithm \ref{wlr} makes Algorithm \ref{bg_wlr} and \ref{inwlr} robust to the outliers for the background modeling problems. Our empirical validation shows that a properly weighted Frobenius norm can perform as well as or better than that of state-of-the-art $\ell_1$ norm minimization (or any other norms) algorithms for background estimation.  

To conclude the paper, we present extensive numerical experiments to show that  our batch-incremental model is as fast as incPCP and ReProCS, also, it can deal with high quality video
sequences as accurately as incPCP and ReProCS. Some conventional
algorithms, such as supervised GFL or ReProCS require
an initial training sequence which does not contain
any foreground object. Our experimental results for both
synthetic and real video sequences show that unlike the supervised
GFL and ReProCS, our models does not require a
prior. We believe that the adaptive nature of our algorithm
is well suited for real time high-definition video
surveillance. 
% and for panning motions of the camera where
%the background slowly evolves.

\vspace{-0.05in}

\section{Weighted low-rank approximation: Theory}
  
\IEEEPARstart{A}s in the standard low rank approximation, the constrained low rank approximation problem \eqref{golub's problem} of Golub, Hoffman, and Stewart has a closed form solution which is given in the following theorem:
\begin{theorem}
\label{theorem 1}\cite{golub}
 Assume ${\rm rank}(A_1)=k$ and $r\ge k$, the solution $\tilde{A}_2$ in~(\ref{golub's problem}) is given by
 \begin{align}\label{ghs}
\tilde{A}_2= P_{A_1}(A_2)+H_{r-k}\left(P^{\perp}_{A_1}(A_2)\right),
 \end{align}
where $P_{A_1}$ and $P^\perp_{A_1}$ are the projection operators to the column space of $A_1$ and its orthogonal complement, respectively.
\end{theorem} 

%\section{Limiting behavior of solutions as $(W_1)_{i,j}\to\infty$ and $(W_2)_{i,j}\to 1$}
Let $(\tilde{X}_1(W)\;~\tilde{X}_2(W))$ be a solution to~(\ref{hadamard problem}). % Denote the solutions to (\ref{golub's problem}) by $A_G$.
Let $\lambda_j = \displaystyle{\min_{1\le i\le m}(W_1)_{ij}}$ and $\lambda = \displaystyle{\min_{1\le j\le k}\lambda_j}$.
Denote $\mathcal{A}=P_{A_1}^\perp(A_2)$ and $\tilde{\mathcal{A}}=P_{\tilde{X}_1(W)}^\perp(A_2).$ Also denote $s={\rm rank}(\mathcal{A})$ and let the ordered non-zero singular values of $\mathcal{A}$ be  $\sigma_1\ge\sigma_2\ge\cdots\ge\sigma_{s}>0.$
\begin{theorem}\label{theorem 3} Suppose that $\sigma_{r-k}>\sigma_{r-k+1}$. Then
we have, as $\lambda\to\infty$ and $W_2=\mathbbm{1}$, %%Let $W_2=\mathbbm{1}_{m\times (n-k)}$.
 	$$
(\tilde{X}_1(W)\;\;\tilde{X}_2(W))=A_G+\displaystyle{O(\frac{1}{\lambda})},
	$$
	where $A_G=(A_1\;\tilde{A}_2)$ is defined to be the unique solution to~(\ref{golub's problem}).
\end{theorem}
\begin{remark}
{\rm (i) We note that $A_G$ is unique due to the assumption $\sigma_{r-k}>\sigma_{r-k+1}$ \cite{golub}. (ii) The convergence $(\tilde{X}_1(W)\;\;\tilde{X}_2(W))\to A_G$ alone is expected and indeed implied by a general result in \cite{markovosky3}. However, it does not specify a rate of convergence.}
\end{remark}

\vspace{-0.1in}
\subsection{A brief literature review}
As it turns out, (\ref{unconstraint closeness}) can be viewed as a generalized total least squares problem~(GTLS) and can be solved in closed form as a special case of weighted low rank approximation with a rank-one weight matrix by using a SVD of the given matrix $(\lambda A_1\;\;A_2)$~\cite{markovosky,markovosky1}. As a consequence of the closed form solutions, one can verify that the solution to (\ref{golub's problem}) is the limit case of the solutions to (\ref{unconstraint closeness}) as $\lambda \to\infty$. Thus, (\ref{golub's problem}) can be viewed as a special case when ``$\lambda = \infty$''. A careful reader may also note that, problem (\ref{unconstraint closeness}) can be cast as a special case of structured low rank problems with element-wise weights~\cite{markovosky3,markovosky4}.
%More specifically, we observe that (\ref{unconstraint closeness}) is contained in the following more general point-wise weighted low rank approximation problem:
%\begin{eqnarray}\label{hadamard problem}
%	\min_{\substack{X_1,X_2\\{\rm rank}(X_1\;\;X_2)\le r}}\|\left((A_1\;\;A_2)-({X}_1\;\;{X}_2)\right)\odot(W_1\;~W_2)\|_F^2,
%\end{eqnarray}
%for $W_1=\lambda \mathbbm{1}$ and $W_2=\mathbbm{1}$ (a matrix whose entries are equal to 1), where $W\in\mathbb{R}^{m\times n}$ is a weight matrix and $\odot$ denotes the Hadamard product.

The weighted low rank approximation problem was studied first when $W$ is an indicator weight for dealing with the missing data case (\cite{wibergjapan,wiberg}) and then for more general weight in machine learning, collaborative filtering, 2-D filter design, and computer vision~\cite{srebro,srebromaxmatrix,Buchanan,manton,lupeiwang,shpak}.
For example, if SVD is used in quadrantally-symmetric two-dimensional~(2-D) filter design, as explained in~\cite{manton} (see also \cite{lupeiwang,shpak}), it might lead to a degraded construction in some cases as it is not able to discriminate between the important and unimportant components of $X$. To address this problem, a weighted least squares matrix decomposition method was first proposed by Shpak~\cite{shpak}. Following his idea of assigning different weights to discriminate between important and unimportant components of the test matrix, Lu, Pei, and Wang~(\cite{lupeiwang}) designed a numerical procedure to solve (\ref{hadamard problem}) for general weight $(W_1\;~W_2)$.

\begin{remark}\label{remark2}
{\rm There is another formulation of weighted low rank approximation problem defined as in~\cite{manton}:
\begin{align}
\label{manton}
\min_{\substack{{X}\in\mathbb{R}^{m\times n}}}\|A-X\|_{Q}^2,~~{\rm subject~to~}{\rm rank}({X})\le r,
\end{align}
where $Q \in \mathbb{R}^{mn\times mn}$ is a symmetric positive definite weight matrix.~Denote $\|A-X\|_Q^2:={\rm vec}(A-X)^TQ{\rm vec}(A-X)$, where ${\rm vec}(\cdot)$ is an operator which maps the entries of $\mathbb{R}^{m\times n}$ to $\mathbb{R}^{mn\times 1}$. It is easy to see that (\ref{hadamard problem}) is a special case of~(\ref{manton}) with a diagonal $Q$. In this paper, we will not use this more general formulation for simplicity.
}
\end{remark}

\vspace{-0.1in}

%\subsection{How to put weight}
\subsection{Why the problem becomes more difficult to solve when weights present}
Note that, in general, the weighted low rank approximation problem  (\ref{hadamard problem}) does not have a closed form solution, even when $W_2=\mathbbm{1}$~\cite{srebro,manton}.
With $W_2=\mathbbm{1}$, we can write (\ref{hadamard problem}) as
\begin{align*}
\min_{\substack{X_1,X_2\\{\rm rank}(X_1\;\;X_2)\le r}}\left(\|(A_1-X_1)\odot W_1\|_F^2+\|A_2-X_2\|_F^2\right).
\end{align*}
Let ${\rm rank}(X_1)=k$. It is easy to see that any $X_2$ such that ${\rm rank}(X_1\;\;X_2)\le r$ can be given in the form
$$X_2=X_1C+BD,$$ for some arbitrary matrices $B\in\mathbb{R}^{m\times (r-k)},$ $D\in\mathbb{R}^{(r-k)\times (n-k)},$ and $C\in\mathbb{R}^{k\times (n-k)}.$ Hence we need to solve
\begin{align}\label{main problem 2}
\min_{X_1,C,B,D}\left(\|(A_1-X_1)\odot W_1\|_F^2+\|A_2-X_1C-BD\|_F^2\right).
\end{align}
Note that, using a block structure, we can write (\ref{main problem 2}) as (with a special low rank structure):
\begin{align*}
\min_{X_1,C,B,D}\left\{\|\left((A_1\;\;A_2)-(X_1\;\;B)\begin{pmatrix} I_k & C\\0 & D\end{pmatrix}\right)\odot (W_1\;\;\mathbbm{1})\|_F^2\right\},
\end{align*}
which is in a form of the alternating weighted least squares problem in the literature~\cite{srebro, markovosky}. But we will not follow the general algorithm proposed in~\cite{markovosky}~for the following reasons: that (i)~due to the special structure of the weight, our algorithm is more efficient than~\cite{markovosky}~(see Algorithm 3.1,~in p.~42~\cite{markovosky}), that (ii) it allows a detailed convergence analysis which is usually not available in other algorithms proposed in the literature~\cite{srebro,manton,markovosky}, and that (iii) it can handle bigger size matrices as we will demonstrate in the numerical result section.
% One is to verify the rate given by Theorem~\ref{theorem 3} numerically and to gain some insight on the sharpness of the rate~($O(\frac{1}{\lambda})$, as $\lambda\to\infty$); the other one is to demonstrate a fast and simple numerical procedure based on alternating direction method in solving the weighted low-rank approximation problem that also
If $k=0,$ then~(\ref{main problem 2}) reduces to an unweighted rank $r$ factorization of $A_2$ and can be solved as an alternating least squares problem~\cite{MC_hastie,als_hanshom}.

\vspace{-0.1in}
\subsection{Solving the weighted problem}
%\subsubsection{Derive a numerical solution}
Denote $F(X_1,C,B, D)=\|(A_1-X_1)\odot W_1\|_F^2+\|A_2-X_1C-BD\|_F^2$ as the objective function. The function $F$ is minimized by using an alternating strategy~\cite{LinChenMa,Boyd_adm} of minimizing the function with respect to each component iteratively:
\begin{align}\label{update rule}
\left\{\begin{array}{ll}
\displaystyle{(X_1)_{p+1}=\arg\min_{X_1}F(X_1,C_p,B_p,D_p)},\\
\displaystyle{C_{p+1}=\arg\min_{C}F((X_1)_{p+1},C,B_p,D_p)},\\
\displaystyle{B_{p+1}=\arg\min_{B}F((X_1)_{p+1},C_{p+1},B,D_p)},\\
\text{and,}\; \displaystyle{D_{p+1}=\arg\min_{D}F((X_1)_{p+1},C_{p+1},B_{p+1},D)}.
\end{array}\right.
\end{align}
Note that each of the minimizing problem for $X_1, C, B,$ and $D$ can be solved explicitly by looking at the gradients of $F(X_1,C,B,D)$. But finding an update rule for $X_1$ turns out to be more involved than the other three variables due to the interference of the weight $W_1$. We update $X_1$ element wise along each row. Therefore we will use the notation $X_1(i,:)$ to denote the $i$-th row of the matrix $X_1$. The numerical process is described in Algorithm \ref{wlr}.
%\subsubsection{Algorithms}
\begin{algorithm}
\begin{small}
	\SetAlgoLined
	\SetKwInOut{Input}{Input}
	\SetKwInOut{Output}{Output}
	\SetKwInOut{Init}{Initialize}
	\nl\Input{$A=(A_1\;\;A_2) \in\mathbb{R}^{m\times n}$ (the given matrix); $W= (W_1\;\;\mathbbm{1})\in\mathbb{R}^{m\times n}$ (the weight), threshold $\epsilon>0$\;}
	\nl\Init {$(X_1)_0,C_0,B_0,D_0$\;}
	%\BlankLine
	\nl \While{not converged}
	{
		\nl $E_p=A_1\odot W_1\odot W_1+(A_2-B_pD_p)C_p^T$\;
		%\BlankLine
		\nl \For {$i=1:m$}
		{	
			\nl $(X_1(i,:))_{p+1}=(E(i,:))_p({\rm diag}(W_1^2(i,1)$ $W_1^2(i,2)\cdots W_1^2(i,k))+C_pC_p^T)^{-1}$\;
		}
		%\BlankLine
		\nl $C_{p+1}=((X_1)_{p+1}^T(X_1)_{p+1})^{-1}(X_1)_{p+1}^T(A_2-B_pD_p)$\;
		\nl $B_{p+1}=(A_2-(X_1)_{p+1}C_{p+1})D_p^T(D_pD_p^T)^{-1}$\;
		\nl $D_{p+1}=(B_{p+1}^TB_{p+1})^{-1}B_{p+1}^T(A_2-(X_1)_{p+1}C_{p+1})$\;
		\nl $p=p+1$\;
	}
	%\BlankLine
	\nl \Output{$(X_1)_{p+1}, (X_1)_{p+1}C_{p+1}+B_{p+1}D_{p+1}.$}
	\caption{WLR Algorithm}\label{wlr}
	\end{small}
\end{algorithm}
\vspace{-0.1in}

\subsection{Convergence}
In this section we comment on the convergence on Algorithm \ref{wlr}. We quote the convergence results of Algorithm \ref{wlr} without their proof. For detailed proofs we refer to \cite{duttali}. The first theorem shows a relation that involves the iterates and the reconstruction error of the objective function. 
\begin{theorem}\label{theorem 7}
	For a fixed $(W_1)_{ij}>0$ let $m_p=F((X_1)_p,C_p,B_p,D_p)$ for $p=1,2,\cdots$ Then,\begin{eqnarray}\label{mj equality}
	m_p-m_{p+1}=&\|((X_1)_p-(X_1)_{p+1})\odot W_1\|^2_F\nonumber\\&+\|((X_1)_p-(X_1)_{p+1})C_p\|^2_F\nonumber\\&+\|(X_1)_{p+1}(C_p-C_{p+1})\|^2_F\nonumber\\
	&+\|(B_p-B_{p+1})D_p\|^2_F\nonumber\\&+\|B_{p+1}(D_p-D_{p+1})\|^2_F.
	\end{eqnarray}
\end{theorem}

\begin{remark}\label{m_p}
From Theorem~\ref{theorem 7} we know that the non-negative sequence $\{m_p\}$ is non-increasing.~Therefore, $\{m_p\}$ has a limit. 
\end{remark}

By using Theorem~\ref{theorem 7} we have the following estimates.
\noindent\begin{corollary}\label{corollary 2}
	We have \begin{enumerate}[(i)]
		\item $m_p- m_{p+1}\geq \frac{1}{2}\|B_{p+1}D_{p+1}-B_pD_p\|_F^2$ for all $p.$
		\item $m_p- m_{p+1}\geq\|((X_1)_p-(X_1)_{p+1})\odot W_1\|^2_F$ for all $p$.
	\end{enumerate}
\end{corollary}

Consider the situation when
\begin{align}\label{mj inequality}
\sum_{p=1}^{\infty}\sqrt{m_p- m_{p+1}}< +\infty.
\end{align}

\begin{theorem}\label{theorem 8}
	\begin{enumerate}[(i)]
		\item  We have the following:~$\sum_{p=1}^{\infty}\|B_{p+1}D_{p+1}-B_pD_p\|_F^2<\infty$, and $\displaystyle{\sum_{p=1}^{\infty}\left(\|((X_1)_p-(X_1)_{p+1})\odot W_1\|_F^2\right)<\infty}$.
		\item If \eqref{mj inequality} holds then $\displaystyle{\lim_{p\to\infty}B_pD_p}$ and $\displaystyle{\lim_{p\to\infty}(X_1)_p}$ exist. Furthermore if we write $L^*:=\displaystyle{\lim_{p\to\infty}B_pD_p}$ then $\displaystyle{\lim_{p\to\infty}B_{p+1}D_{p}=L^*}$ for all $p.$
	\end{enumerate}
\end{theorem}

Note that Corollary~\ref{theorem 8} only states the convergence of the sequence $\{B_pD_p\}$ but not of $\{B_p\}$ and $\{D_p\}$ separately. The convergence of $\{B_p\}$ and $\{D_p\}$ can be obtained with stronger assumption as demonstrated in the next result. 
\begin{theorem} \label{theorem 9}
	Assume~(\ref{mj inequality}) holds.
	\begin{enumerate}[(i)]
		\item If $B_p$ is of full rank and $B_p^TB_p\ge\gamma I_{r-k}$ for large $p$ and some $\gamma>0$ then $\displaystyle{\lim_{p\to\infty}D_p}$ exists.
		\item If $D_p$ is of full rank and $D_pD_p^T\ge\delta I_{r-k}$ for large $p$ and some $\delta>0$  then $\displaystyle{\lim_{p\to\infty}B_p}$ exists.
		\item If $X_1^*:=\displaystyle{\lim_{p\to\infty}(X_1)_{p}}$ is of full rank, then $C^*:=\displaystyle{\lim_{p\to\infty}C_p}$ exists. Furthermore, if we write $L^*=B^*D^*,$ for $B^*\in\mathbb{R}^{m\times(r-k)}, D^*\in\mathbb{R}^{(r-k)\times(n-k)}$, then $({X}_1^*,C^*,B^*,D^*)$ will be a stationary point of $F$.
	\end{enumerate}
\end{theorem}

\begin{figure*}
    \centering
    \includegraphics[width=\textwidth, height = 2.6in]{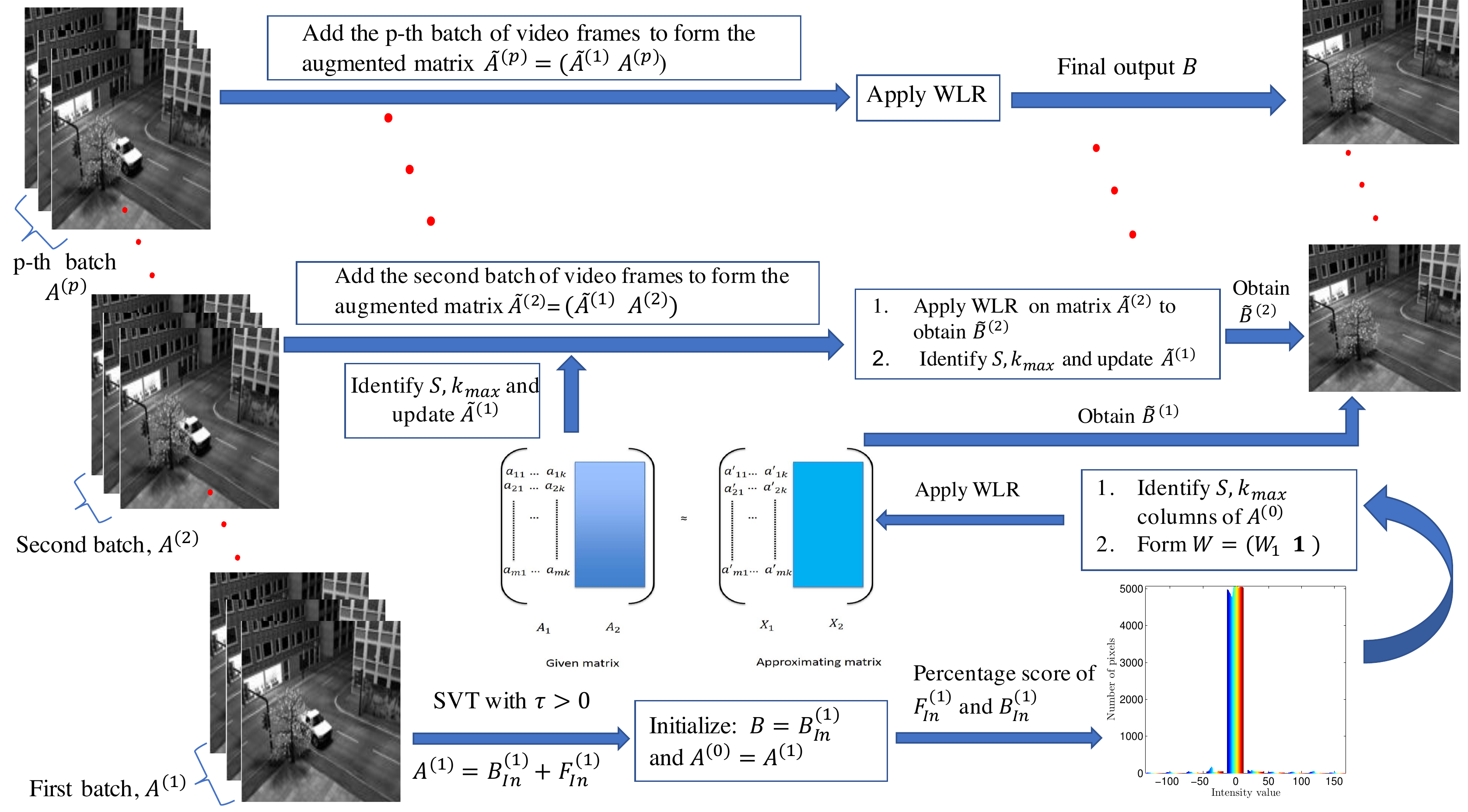}
    \caption{\small{A flowchart for WLR inspired batch-incremental background estimation model proposed in Algorithm \ref{inwlr}.}}
    \label{algo_flowchart}
\end{figure*}

\vspace{-0.1in}

\section{Weighted low-rank approximation: application to background modeling}
\IEEEPARstart{W}e focus on two algorithms for background modeling by using Algorithm \ref{wlr}. Both algorithms robustly determine the frame indices which are contenders of the background frames (with least or no foreground movement) and then take the advantage of weight matrix $W$ to estimate the background and the foreground. 
\begin{algorithm}\begin{small}
	\SetAlgoLined
	\SetKwInOut{Input}{Input}
	\SetKwInOut{Output}{Output}
	\SetKwInOut{Init}{Initialize}
	\nl\Input{$A=(A_1\;\;A_2) \in\mathbb{R}^{m\times n}$ (the given matrix); $W= (W_1\;\;\mathbbm{1})\in\mathbb{R}^{m\times n}$,(the weight), threshold $\epsilon>0,$ $i_1,i_2\in\mathbb{N}$\;}
	\nl Run PCA to get low rank $B_{In}$ and $F_{In}=A-B_{In}$\;
	\nl Learn background frame indices $S$ from $B_{In}$ and $F_{In}$\;
	%Use the histogram of $F_{In}$ to define the threshold $\epsilon_1$\;
	%set $F_{In}(F_{In}\le\epsilon_1)=0$ and $F_{In}(F_{In}>\epsilon_1)=1$\; 
	%\nl Convert $B_{In}$ directly to a logical matrix $LB_{In}$\;
	%\nl Find
	%$\epsilon_2={\rm mode}(\{\frac{\sum_i(LF_{IN})_{i1}}{\sum_i(LB_{IN})_{i1}},\frac{\sum_i(LF_{IN})_{i2}}{\sum_i(LB_{IN})_{i2}},\cdots, \frac{\sum_i(LF_{IN})_{in}}{\sum_i(LB_{IN})_{in}}\})
	%$\;
	%\nl Denote $S = \{i:(\frac{\sum_i(LF_{IN})_{i1}}{\sum_i(LB_{IN})_{i1}}\le \epsilon_2\}$\;
	\nl Set $k=\Bigl\lceil|S|/i_1\Bigr\rceil, r = k+i_2$\;
	\nl Rearrange data:~$\tilde{A}_1= (A(:,i))_{m\times k}$, randomly chosen $k$ frames from $i\in S$ and $\tilde{A}_2= (A(:,i'))_{m\times (n-k)}$, $i'$ from the remaining frames\;
	\nl Apply WLR on $\tilde{A}=(\tilde{A}_1\;\tilde{A}_2)$ to obtain $\tilde{X}$\;
	\nl Rearrange the columns of $\tilde{X}$ similar to $A$ to find $X$\;
	\nl \Output{$X$.}
	\caption{Background Estimation using WLR}\label{bg_wlr}
	\end{small}
\end{algorithm}

\begin{algorithm}
\begin{small}
    \SetAlgoLined
    \SetKwInOut{Input}{Input}
    \SetKwInOut{Output}{Output}
    \SetKwInOut{Init}{Initialize}
    \nl\Input{$p$,~$A=(A^{(1)}\;\;A^{(2)}\ldots A^{(p)}) \in\mathbb{R}^{m\times n}$,~$\tau>0$~(for SVT), $\alpha, \beta>0$~(for weights),
    threshold $\epsilon>0,$ $k_{\rm max},~i_r\in\mathbb{N}$\;}
    \nl Run SVT on $A^{(1)}$ with parameter $\tau$ to obtain: $A^{(1)}=B^{(1)}_{In}+F^{(1)}_{In}$\;
    \nl Initialize the background block by $B=B^{(1)}_{In}$ and $A^{(0)}=A^{(1)}$\;
    \nl \For {$j=1:p$}
    {
    \nl Identify the indices $S$ of at most $k_{max}$ columns of $A^{(j-1)}$ that are closest to background using $B$ and
    $F=A^{(j-1)}-B$\;
    \nl Set $k=\#(S), r = k+i_r$\;
    \nl Set the first block:~$\tilde{A}_1= (A^{(j-1)}(:,i))_{m\times k}$ with $i\in S$\;
    \nl Define $W=(W_1\;\mathbbm{1})$ with $W_1\in\mathbb{R}^{m\times k}$ where $(W_1)_{ij}$ are randomly chosen
    from $[\alpha, \beta]$\;
    \nl Apply {\bf Algorithm 1} on $\tilde{A}^{(j)}=(\tilde{A}_1\;A^{(j)})$ using threshold $\epsilon$ and weight $W$ to obtain its
    low rank component $\tilde{B}^{(j)}$ and define $\tilde{F}^{(j)}=\tilde{A}^{(j)}-\tilde{B}^{(j)}$\;
    \nl Take the sub-matrix of $\tilde{B}^{(j)}$ corresponding to the $A^{(j)}$ block such that $A^{(j)}=B^{(j)}+F^{(j)}$\;
    \nl Update the background block: $B=\tilde{B}^{(j)}$\;
    }
    \nl \Output{$B=(B^{(1)},B^{(2)},...,B^{(p)})$.}
    \caption{Incremental Background Estimation using WLR~(inWLR)}\label{inwlr}
    \end{small}
\end{algorithm}
\begin{figure}
    \centering
    \includegraphics[width=.4\textwidth]{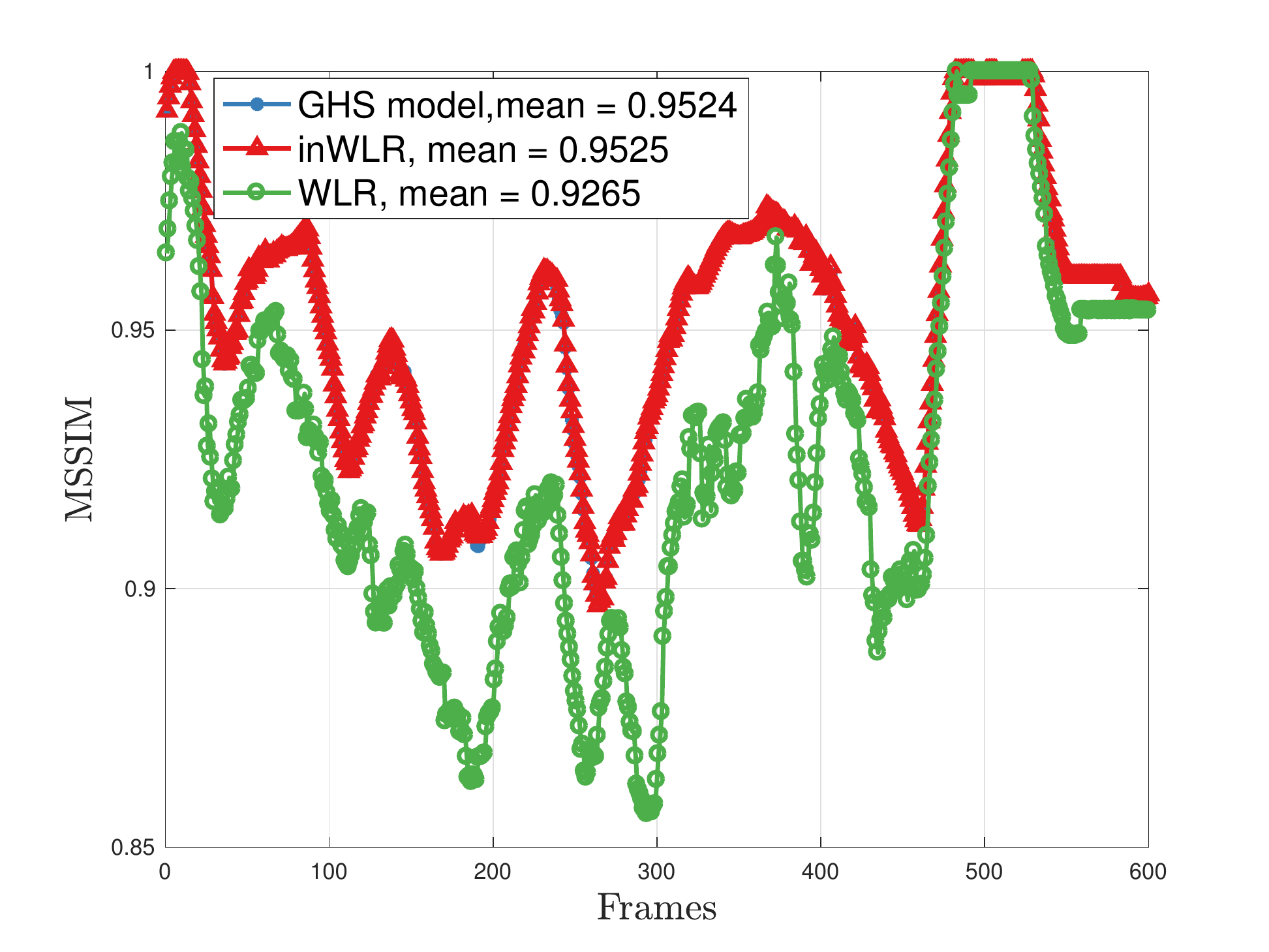}
    \caption{\small{Comparison of MSSIM of WLR acting on all frames,~inWLR ($p=6$),~and GHS inspired background estimation model with resolution $[144,176]$.}}
    \label{ssim}
\end{figure}

\subsection{Background estimation by using Algorithm \ref{bg_wlr}}
First, we describe a batch background modeling
algorithm. To initialize, we first solve WLR (by using Algorithm \ref{wlr}) for $W =I_n$
(this is just PCA) to obtain an coarse estimate 
of the background and foreground: $A = B_{In}+F_{In}$,
where $B_{In}$ is a low rank approximation to $A$ given by PCA.
Next, we use $B_{In}$ and $F_{In}$ to learn the frame indices that
are closest to the pure background (with least or no foreground movements). This is done heuristically (as in \cite{duttaligongshah}(see Figure 2)). 
By setting a threshold $\epsilon_1 > 0$ based on
the histogram of $F_{In}$, we convert $F_{In}$ into a binary matrix
$LF_{In}$: all entries of $F_{In}$ that are bigger than $\epsilon_1$ are replaced by 1
and the others are replaced by 0. The matrix $B_{In}$ is directly
converted to a binary matrix $LB_{In}$. Next, we calculate the
ratios of the frame sum (i.e. the column sum) of $LF_{In}$ to the
corresponding frame sum of $LB_{In}$ and identify the indices
with ratios less than the mode of these ratios as possible
pure background frame indices. Finally, we apply WLR by
putting the weight at the learned frame indices to decompose
the data matrix $A$ into background and foreground:
$A = B + F$. Our experiments show that the performance
depends more on the correct location (indices) of the background
frames than on the values of the weight. We remark
that Dutta and Li \cite{duttali_acl} and Xin et al. \cite{xin2015} used the pure
background frames in their background estimation model,
assuming the frames were already given. On the contrary,
Algorithm 2 learns the background frame indices from the
data, thus providing a robust background estimation model.

%\vspace{-0.05in}
\subsection{An incremental model using WLR}
Next we propose an incremental weighted
low rank approximation~(inWLR) algorithm for background modeling~(see Algorithm \ref{inwlr} and Figure~\ref{algo_flowchart}). 
This algorithm proves to be efficient compared to Algorithm \ref{bg_wlr} when the video sequence contains a lot of frames. 
Our algorithm fully exploits WLR, in which a prior knowledge
of the background space can be used as an additional constraint to
obtain the low rank (thus the background) estimation of the data
matrix $A$. First, we partition the original video
sequence into $p$ batches:~$A=(A^{(1)}\;\;A^{(2)}\ldots A^{(p)})$,
where the batch sizes do not need to be equal.~Instead of working on
the entire video sequence, the algorithm incrementally works through
each batch. To initialize, the algorithm coarsely
estimates the possible background frame indices of $A^{(1)}$: we
run the classic singular value thresholding~(SVT) of Cai
et. al. ~\cite{caicandesshen} on $A^{(1)}$ (we can afford this since the size of $A^{(1)}$ is much smaller than that of $A$) to obtain a low rank
component (containing the estimations of background frames)
$B^{(1)}_{In}$ and let $F^{(1)}_{In}=A^{(1)}-B^{(1)}_{In}$ be the
estimation of the foreground matrix (Step 2).~From the above estimates, we
obtain the initialization for $B$ and $A^{(0)}$~(Step 3).~Then, we
go through each batch $A^{(j)}$, using the estimates of the
background from the previous batch as prior for the WLR algorithm to obtain the background $\tilde{B}^{(j)}$~(Step 9).~To
determine the indices of the frames that contain the least information of
the foreground we identify the ``best background frames'' by using a modified version of
the percentage score model by Dutta et. al. ~\cite{duttaligongshah}~(Step 5).~Using this modified model allows us to estimate $k$, $r$, and
the first block $\tilde{A}_1$ which contains the background prior
knowledge (Steps 6-7).~Weight matrix $W= (W_1\;\mathbbm{1})$ is
chosen by randomly picking the entries of the first block $W_1$ from an
interval $[\alpha, \beta]$ by using an uniform distribution, where
$\beta>\alpha>0$ are large (Step 8).~To understand the effect of
using a large weight in $W_1$, we refer the reader
to Theorem~\ref{theorem 3} and \cite{duttali_acl,duttali}.~Finally, we collect background
information for next iteration~(Steps 10-11).~Note that the number
of columns of the weight matrix $W_1$ is $k$, which is controlled by
bound $k_{max}$ so that the column size of $\tilde{A}^{(j)}$ does not
grow with $j$.~The output of the algorithm is the background
estimations for each batch collected in a single matrix $B$.~When
the camera motion is small, updating the first block matrix
$\tilde{A}_1$ (Step~7) has trivial impact on the model since it does not change much. However, when the %camera is panning and the
background is continuously evolving (but slowly),~our inWLR could be proven very robust as new frames are entering in the video. Moreover, we show that Algorithm \ref{inwlr} is faster compare with Algorithm \ref{bg_wlr}.
\begin{figure}
    \centering
    \includegraphics[width=0.5\textwidth]{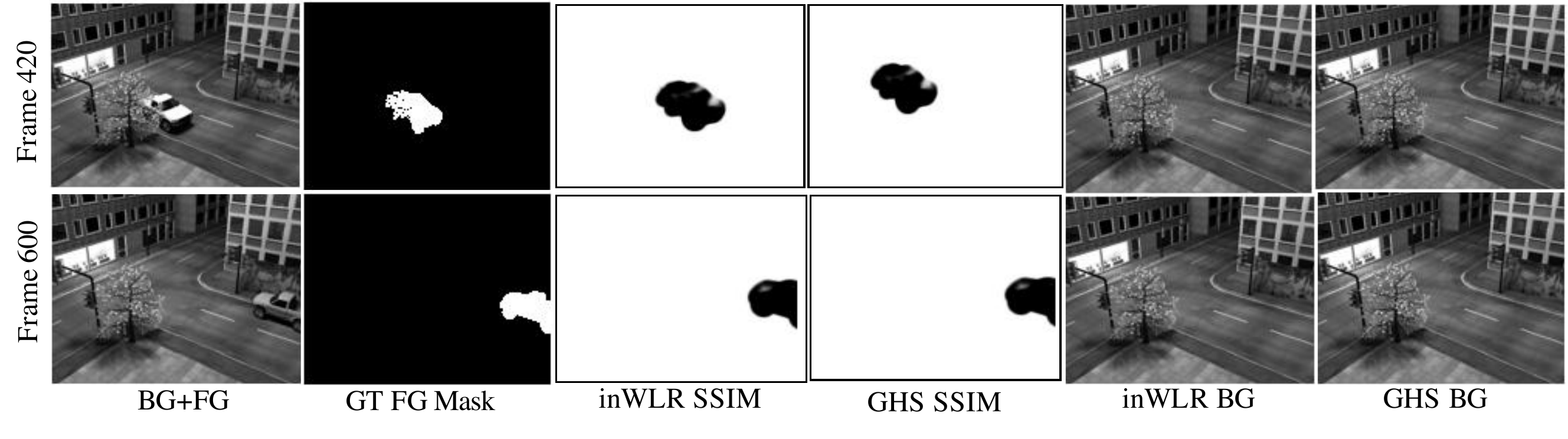}
    \caption{\small{SSIM map of inWLR and GHS inspired background estimation model.Top to bottom: Frame 420 with dynamic foreground, frame 600 with static foreground. SSIM index of the methods are 0.95027 and 0.96152, respectively.}}
    \label{ssim_map}
\end{figure}
\begin{figure}
    \centering
    \includegraphics[width = 0.4\textwidth]{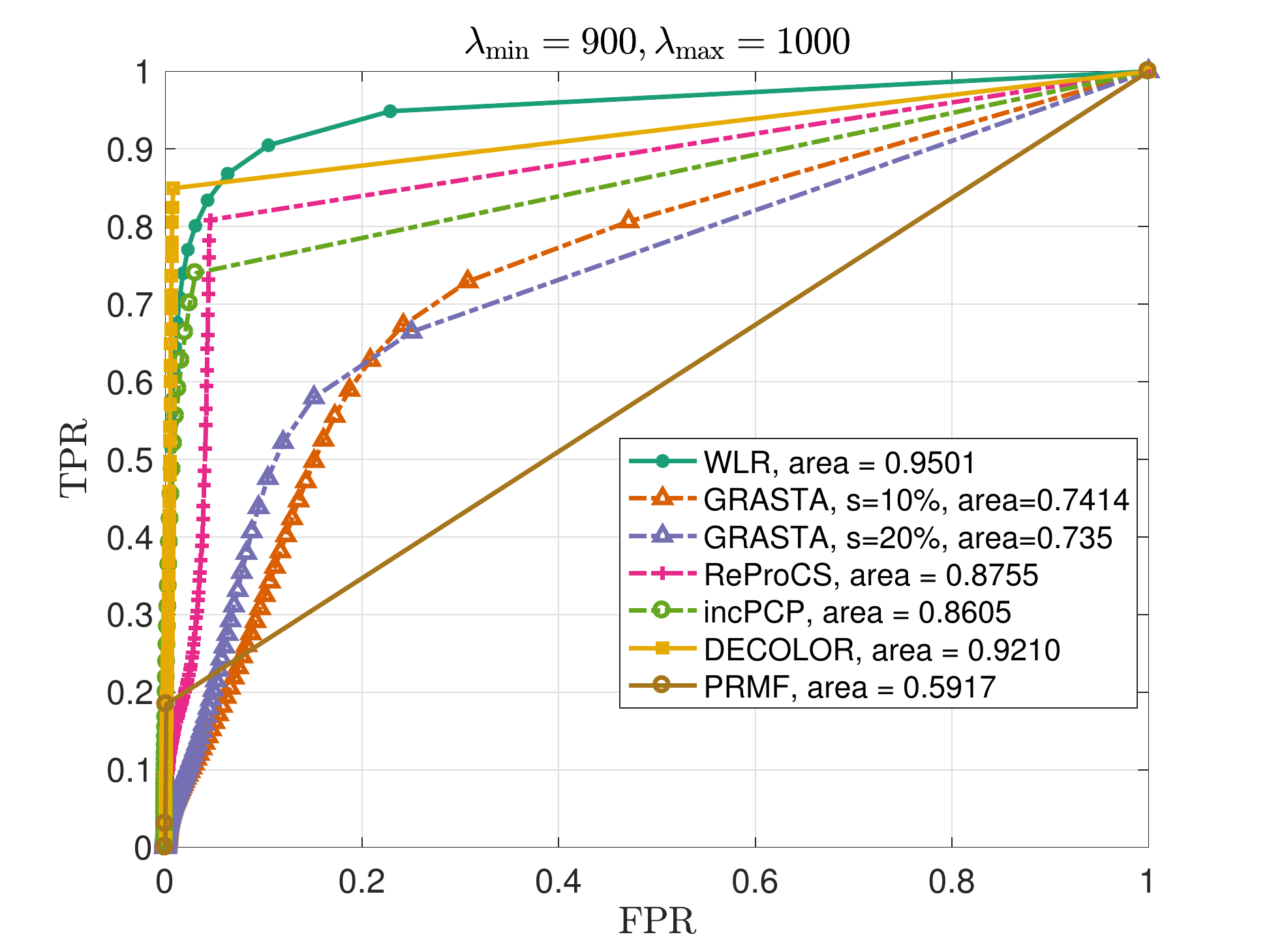}
   \caption{\small{ROC curve to compare between WLR, GRASTA, ReProCS, incPCP, DECOLOR, and PRMF on {\tt Basic} video, resolution $144\times176$.}}
\label{roc_wlr}
\end{figure}
\begin{figure}
    \centering
    \includegraphics[width = 0.4\textwidth]{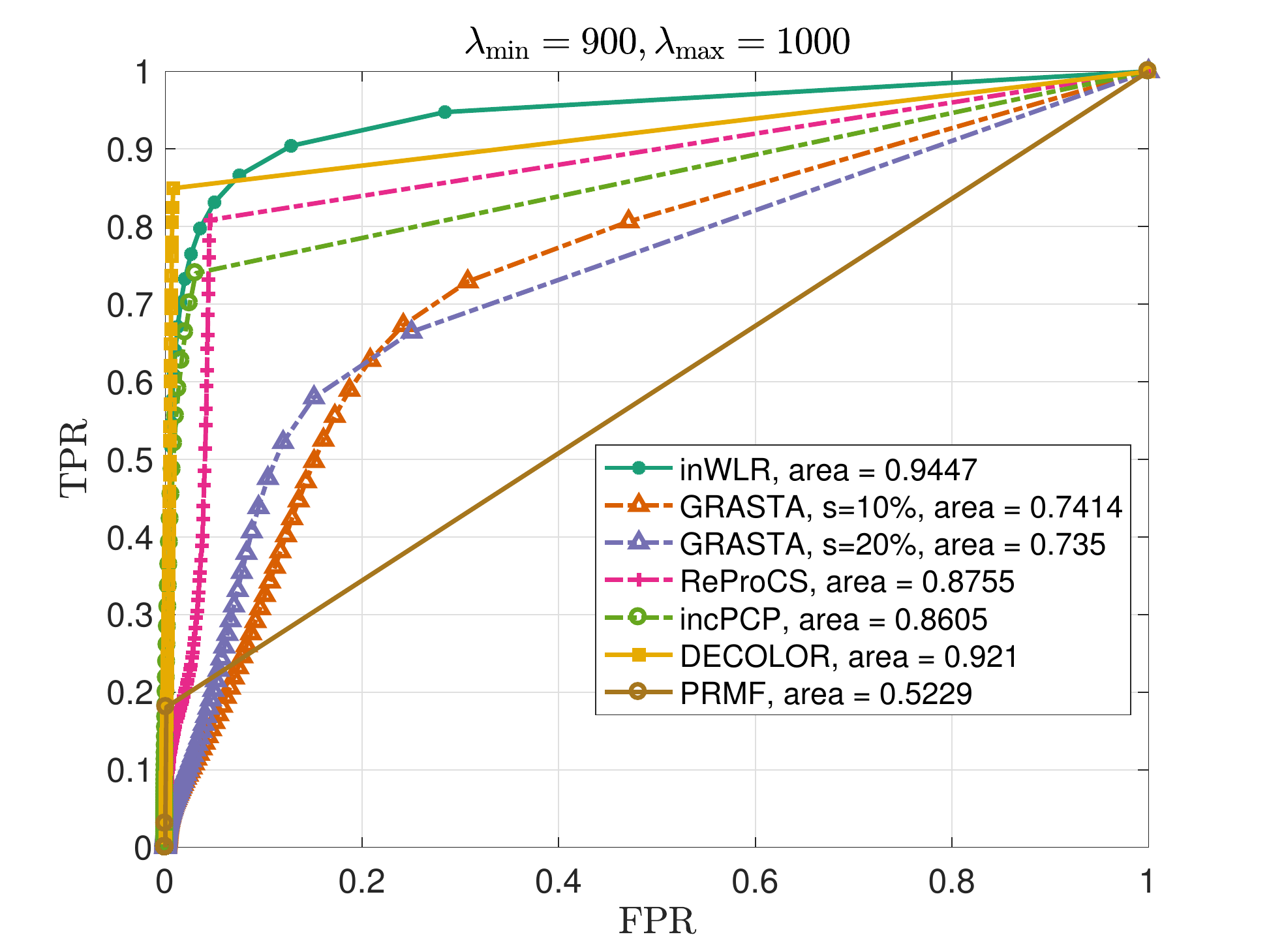}
   \caption{\small{ROC curve to compare between inWLR, GRASTA, ReProCS, incPCP, DECOLOR, and PRMF on {\tt Basic} video, resolution $144\times176$.}}
\label{roc_inwlr}
\end{figure}

%\vspace{-0.1in}

\subsection{Complexity Analysis}
Now, we analyze the complexity of
Algorithm~\ref{inwlr} for equal batch size and compare it with Algorithm \ref{bg_wlr}. Primarily, the cost of
the SVT algorithm in Step~2 is only
$\mathcal{O}(\frac{mn^2}{p^2})$.~Next, in Step~9, the complexity of
implementing Algorithm~\ref{wlr}
is~$\mathcal{O}(mk^3+\frac{mnr}{p})$. Note that $r$ and $k$ are
linearly related and $k \leq k_{\rm max}$.~Once we obtain a refined
estimate of the background frame indices $S$ as in Step~5 and form
an augmented matrix by adding the next batch of video frames, a very
natural question in proposing our WLR inspired~Algorithm~\ref{inwlr}
is:~why do we use Algorithm~\ref{wlr} in each incremental
step~(Step~9) of Algorithm~\ref{inwlr} instead of using a closed
form solution (\ref{ghs}) of GHS? We justify as follows:~the estimated background frames $\tilde{A}_1$ are not
necessarily exact background; they are only estimations of background.~Thus,
GHS inspired model may be forced to follow the wrong data while
inWLR allows enough flexibility to find the best fit to the
background subspace. This is confirmed by our numerical experiments
(see Section~\ref{experiemnts} and Figure~\ref{ssim}). Thus, to analyze the
entire sequence in $p$ batches, the complexity of
Algorithm~\ref{inwlr} is approximately
$\mathcal{O}(m(k^3p+nr))$.~Note that the complexity of
Algorithm~\ref{inwlr} is dependent on the partition $p$ of the
original data matrix. Our numerical experiments suggest that for video
frames of varying sizes, the choice of $p$ plays an important role
and is empirically determined.
Unlike~Algorithm~\ref{inwlr}, if Algorithm~\ref{bg_wlr} is used on the
entire data set and if the number of possible background frame indices
is $k'$, then the complexity is $\mathcal{O}(m{k'}^3+mnk')$. When
$k'$ grows with $n$ and becomes much bigger than $k_{max}$ in order to achieve competitive performance, we see that Algorithm~\ref{wlr} tends to slow down with higher
overhead than Algorithm~\ref{inwlr} does. ~%(see Table~\ref{time}).
\begin{figure*}
    \centering
    \includegraphics[width=\textwidth]{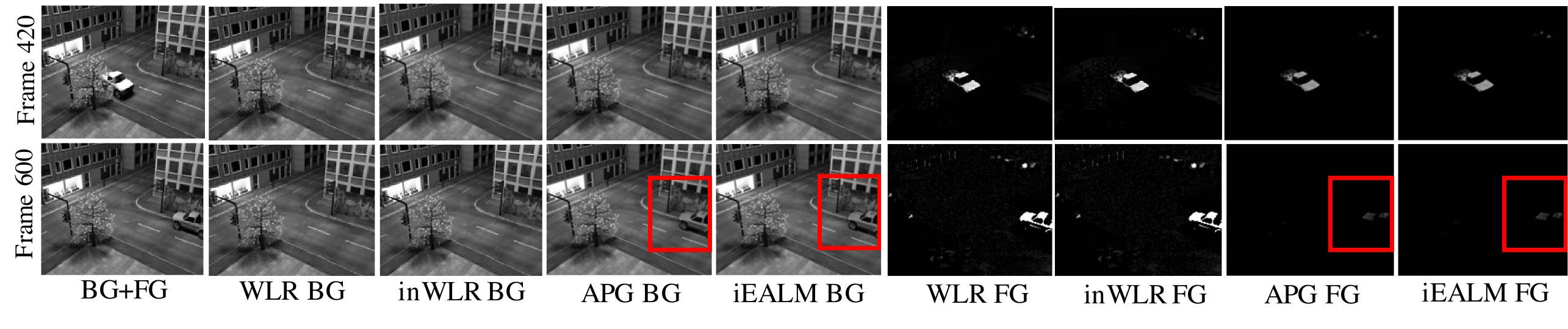}
    \caption{\small{Qualitative comparison of WLR and inWLR with iEALM and APG. Top to bottom: Frame 420 with dynamic foreground, frame 600 with static foreground.}}
    \label{rpca}
\end{figure*}

%\vspace{-0.1in}

\section{Experiments}\label{experiemnts}
\IEEEPARstart{W}e present our extensive numerical experiments and show the effectiveness of our background modelling algorithms on synthetic and real world video sequences and compare them with several state-of-the-art background modeling algorithms. 
\subsection{Data sets}
%We start with a brief overview of the video data-sets used in this paper. 
We extensively use 24 gray scale videos from the Stuttgart, I2R, Wallflower, CDNet 2014, and the SBI dataset~\cite{cvpr11brutzer,lidata,sbi_a,SBI_web,wallflower,cdnet}. Stuttgart is a synthetic dataset, but all other datasets are real world videos. They contain several challenges, for example, static foreground, dynamic background, change in illumination, occlusion and disocclusion of static and dynamic foreground. We refer the readers to Table \ref{dataset} to get an overall idea of the number of frames of each video sequence used, video type, and resolution.

\begin{table}
\footnotesize
\begin{center}
\begin{tabular}{|l|c|c|c|}
\hline
Dataset & Video & Frames & Resolution\\
&&& used\\
\hline
Stuttgart~\cite{cvpr11brutzer} & {\tt Basic} & 600 & $144\times176$  \\
Wallflower~\cite{wallflower} &{\tt Waving Tree} &66 & $120\times 160$ \\
%\hline
SBI~\cite{sbi_a}& {\tt Snellen} &321& $144\times176$ \\
& {\tt IBMTest2} &90& $144\times144$ \\
& {\tt HumanBody} & 740 & $144\times144$ \\
& {\tt Foliage} & 300& $144\times 176$ \\
& {\tt Candela} &350 & $144\times176$\\
& {\tt Caviar2} &460 &$144\times176$  \\
& {\tt Hall and Monitor} & 296 & $144\times176$ \\
CDnet 2014~\cite{cdnet}& {\tt Abandoned Box} & 300& $144\times176$ \\
& {\tt Backdoor} &300 & $144\times176$\\
& {\tt Busstation} &600 &$144\times176$  \\
& {\tt Intermittent Pan} &400 & $144\times176$ \\
%\hline
& {\tt Tunnel Exit} & 300& $144\times 176$ \\
&{\tt Port} &1000 & $144\times176$\\
&{\tt Fountain} &300 & $144\times176$ \\
&{\tt Overpass} & 1100& $144\times176$ \\
&{\tt Fountain 2} &1100 & $144\times176$\\
&{\tt Canoe} &1100 & $144\times176$\\
&{\tt Fall} &1100 & $144\times176$\\

I2R/Li dataset~\cite{lidata} & {\tt Meeting Room} &1209 & $64\times 80$ \\
&{\tt Watersurface} &162 & $128\times 160$ \\
&{\tt Campus} & 600 & $64\times 80$ \\
&{\tt Fountain} & 500 & $64\times 80$ \\
\hline
\end{tabular}
\end{center}
\caption{\small{Data used in this paper.}}\label{dataset}
\end{table}
\begin{table*}
\scriptsize
\begin{center}
\begin{tabular}{|l|c|c|c|c|}
\hline
\;\;Algorithm & Appearing in  & Reference\\
\hline
Weighted Low-rank approximation (WLR)  &Figure \ref{ssim}, \ref{roc_wlr}, \ref{rpca}, \ref{msim_wlr}, \ref{420st}, and \ref{dyn_for}& This paper (Algorithm \ref{bg_wlr}), \cite{duttali_bg} \\
Incremental Weighted Low-rank approximation (inWLR)  & Figure \ref{algo_flowchart}, \ref{ssim},  Figure \ref{roc_inwlr}-\ref{CDnet_result1}, Figure \ref{WT_32}-\ref{CDnet_result2}, Table \ref{sbi_data}, and \ref{cdnet_data} & This paper (Algorithm \ref{inwlr}), \cite{inWLR}\\
Inexact Augmented Lagrange Method of Multipliers (iEALM) &Figure \ref{rpca}&\cite{LinChenMa}  \\
Accelerated Proximal Gradient (APG) &Figure \ref{rpca} &\cite{APG}  \\
Goulb et. al. inspired BG model & Figure \ref{ssim}, \ref{ssim_map} & \cite{golub,duttali_bg}\\
%\hline
Supervised Generalized Fused Lasso (GFL) & Figure \ref{600st}, \ref{WT_32}& \cite{xin2015} \\
%\hline
 Grassmannian Robust Adaptive Subspace
Tracking (GRASTA) &Figure~\ref{roc_wlr}, \ref{roc_inwlr}, \ref{420st}  & \cite{grasta} \\
%\hline
Recurssive Projected Compressive Sensing (ReProCS) &Figure \ref{roc_wlr}, \ref{roc_inwlr}, \ref{msim_wlr}, \ref{123st}, \ref{WT_32} &\cite{reprocs,pracreprocs,modified_cs}  \\
%\hline
Incremental Principal Component Pursuit (incPCP) & Figure \ref{roc_wlr}, \ref{roc_inwlr}, \ref{msim_wlr}, \ref{420st}, \ref{600st} &\cite{incpcp,matlab_pcp,inpcp_jitter}\\
Probabilistic Robust Matrix Factorization (PRMF) & Figure \ref{roc_wlr}, \ref{roc_inwlr}, and \ref{600st} & \cite{prmf}\\
Detecting Contiguous Outliers in the Low-Rank Representation (DECOLOR) & Figure \ref{roc_wlr}, \ref{roc_inwlr}, and \ref{600st} & \cite{decolor}\\
\hline
\end{tabular}
\end{center}
\caption{\small{Algorithms compared in this paper.}}\label{algo}
\end{table*}

\vspace{-0.1in}

\subsection{Metrics used for quantitative comparison}
We use four different metrics for this purpose: traditionally
used receiver and operating characteristic (ROC) curve, peak signal to noise ratio (PSNR), and the most advanced
measures mean structural similarity index (MSSIM), and multiscale structural similarity index (MSSSIM)\cite{mssim,msssim, boumans_taxonomy} as they mostly agree with the human visual perception \cite{mssim}. When a ground truth mask (foreground or background) is available for each video frame,
we use a pixel-based measure of $F$ or $B$, the foreground or background recovered
by each method to form the confusion matrix for the ROC
predictive analysis. In our case, the pixels are represented
by 8 bits per sample, and $M_I$, the maximum pixel
value of the image is 255. Therefore, a uniform threshold
vector {\tt linspace}$(0, M_I , 100)$ is used to compare the pixelwise
predictive analysis between each recovered foreground or background
frame and the corresponding ground truth. PSNR is calculated by using the metric $10log_{10}\frac{M_I^2}{MSE},$ such that ${\rm MSE}=\frac{1}{mn}\|G(:,i)-R(:,i)\|_2^2$, where $R(;,i)$ is the recovered vectorized BG/FG frame and $G(:,i)$ is the corresponding vectorized GT frame. To calculate the SSIM and MSSSIM of each recovered FG/BG video
frame, we consider a $11\times11$ Gaussian window with standard deviation ($\sigma$) 1.5 unless otherwise specified. We consider the area covered by the ROC curve of an algorithm in a unit square as a measure of its performance, where the higher the value is the better. Similarly, for SSIM and MSSSIM the values that are closer to 1 are better. Lastly, the PSNR of a reconstructed image generally falls in the range 30-50 dB, where the higher the better as well.
\begin{figure}
    \centering
    \includegraphics[width = 0.4\textwidth]{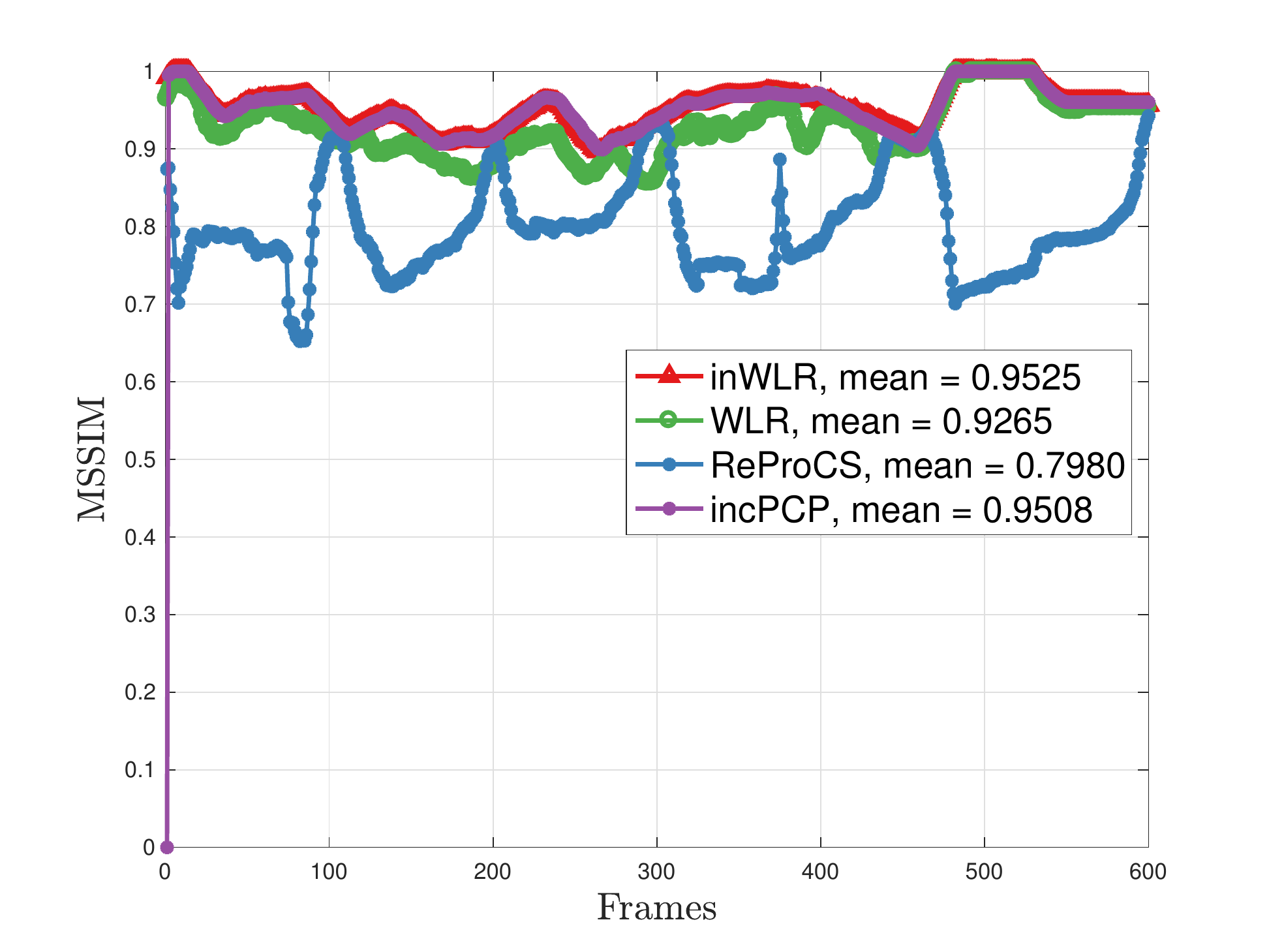}
      \caption{\small{Mean SSIM to compare between WLR, inWLR, ReProCS,  and incPCP on {\tt Basic} video, frame size $144\times176$.}}
\label{msim_wlr}
\end{figure}
\begin{figure}
    \centering
    \includegraphics[width=0.5\textwidth]{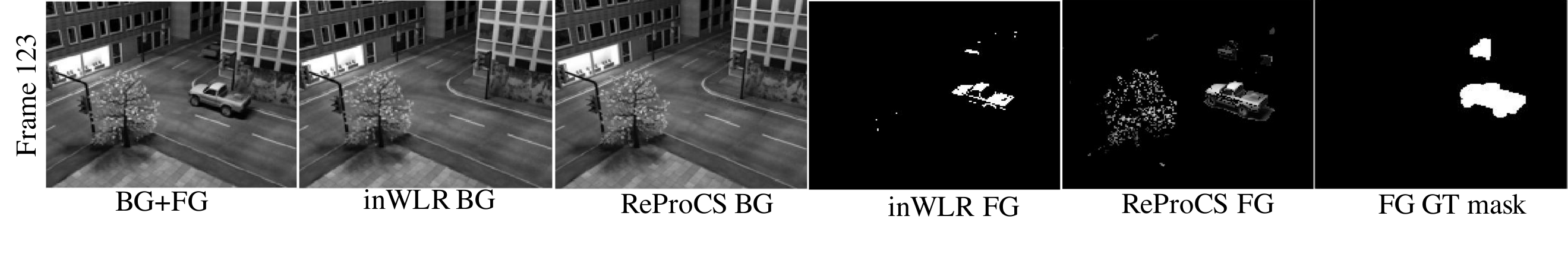}
    \caption{\small{Basic scenario frame 123. Both methods recover similar quality background, however, ReProCS foreground has more false positives than inWLR. This explains why ReProCS suffers in quantitative evaluation.}}
    \label{123st}
\end{figure}

\subsection{Results on {\tt Basic} scene of Stuttgart artificial video}
Due to the availability of ground truth frames for each
foreground mask, we use 600 frames of the {\tt Basic} scenario
of the Stuttgart artificial video sequence \cite{cvpr11brutzer} for
quantitative and qualitative analysis. To capture an unified comparison
against each method, we resize the video frames to
$144\times176$ and for inWLR set $p = 6$; that is, we add a batch
of 100 new video frames in every iteration until all frames
are exhausted.
\begin{figure*}
    \centering
    \includegraphics[width=\textwidth]{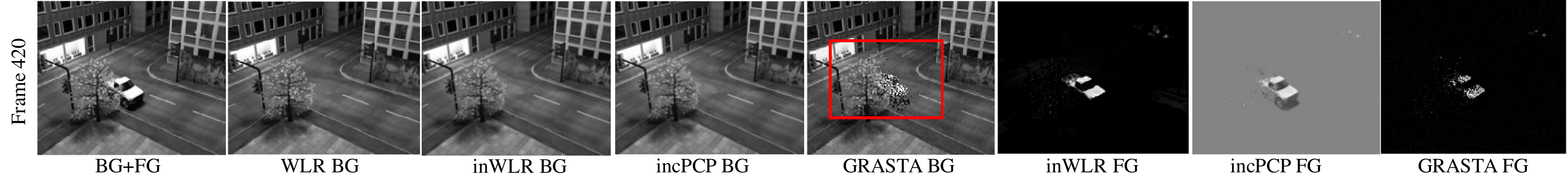}
    \caption{\small{{\tt Basic} scenario frame 420.GRASTA with subsample rate 10\% recovers a fragmentary foreground and degraded background.}}
    \label{420st}
\end{figure*}
\begin{figure*}
    \centering
    \includegraphics[width=\textwidth]{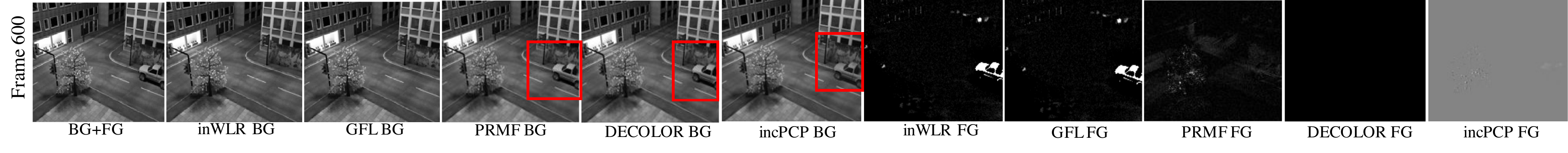}
    \caption{\small{{\tt Basic} scenario frame 600. incPCP, PRMF, DECOLOR fail to detect the static foreground object, though a careful reader can detect a blurry reconstruction of the car in incPCP foreground. However, the SSIM map of all methods are equally good.}}
    \label{600st}
\end{figure*}
\subsubsection{Comparison with GHS}
Because the Basic scenario has no noise, once we estimate
the background frames, GHS can be used as a baseline method for comparing the effectiveness of Algorithm \ref{inwlr}. To
demonstrate the benefit of using an iterative process as inWLR~(Algorithm \ref{wlr}), we first compare the performance of Algorithm \ref{inwlr} against the GHS inspired model. We also compare
Algorithm \ref{bg_wlr} acting on all 600 frames with the parameters
specified in \cite{duttali_bg}. We use MSSIM to quantitatively evaluate the overall image quality. To calculate MSSIM we perceive the information of how
the high-intensity regions of the images come through the
noise, and consequently, we pay much less attention to the
low-intensity regions. We remove the noisy components $F$ by thresholding it by $\epsilon_1$. Figure \ref{ssim} indicates that the inWLR and GHS inspired model produce the same result, but inWLR is more time efficient than GHS. To process 600 frames, inWLR takes \textbf{18.06} seconds. In contrast, GHS inspired model takes 160.17 seconds and WLR takes \textbf{39.99} seconds. In Figure \ref{ssim_map}, the SSIM
map of two sample video frames of the {\tt Basic} scenario show
that both methods recover the similar quality background. 
\begin{figure}
	\centering  
	\includegraphics[width=\linewidth, height = 4.6in]{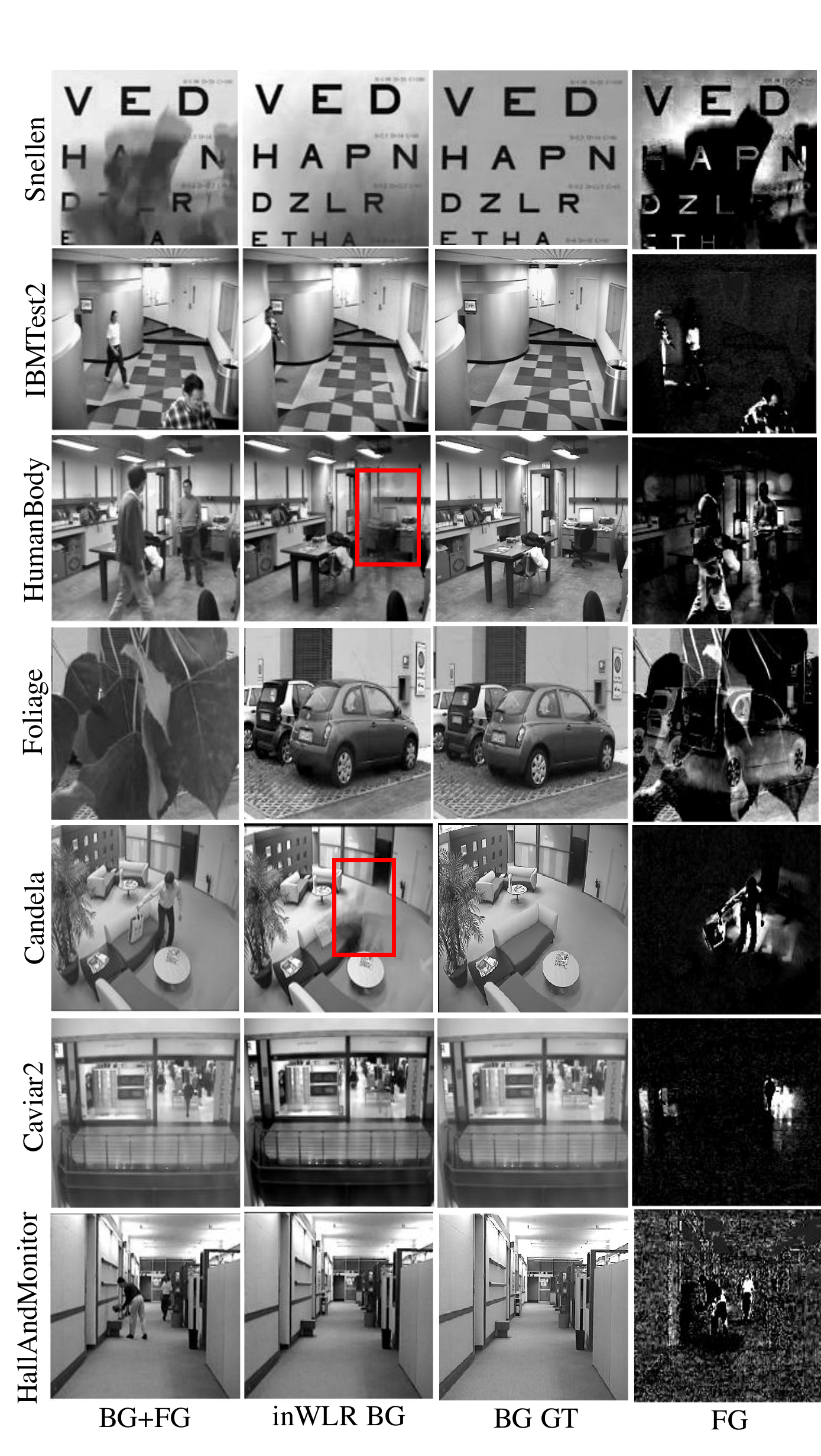}
	\caption{\small{Qualitative results of inWLR on selected sequences of SBI dataset.}}\label{SBI_result}
\end{figure}
\begin{table}
\begin{center}
\begin{tabular}{|l|c|c|c|c|}
\hline
Video & MSSIM & MSSSIM & PSNR & Area under\\&&&& ROC curve \\
\hline
{\tt Snellen} &0.99179 & 0.99891 & 60.9451 & 0.9525\\
\hline
{\tt IBMtest2}& 0.99979 & 0.99998 &76.2776 & 0.9985\\
\hline
{\tt Human Body} &0.9994  &0.9999  &71.5218 &-\\
\hline
{\tt Foliage} &0.9865 & 0.99803 & 58.5055  &-\\
\hline
{\tt Candela} &0.9999 & 0.999995 & 80.2983 &0.9988\\
\hline
{\tt Caviar2} &0.99998 & 0.99999 & 87.8575 & - \\
\hline
{\tt HallMonitor} & 0.99993& 0.99998 &84.8487 &1\\
\hline
\end{tabular}
\end{center}
\caption{\small{Performance of inWLR on SBI dataset~\cite{sbi_a}}.}\label{sbi_data}
\end{table}
\begin{figure}
	\centering  
	\includegraphics[width=\linewidth, height = 4.5in]{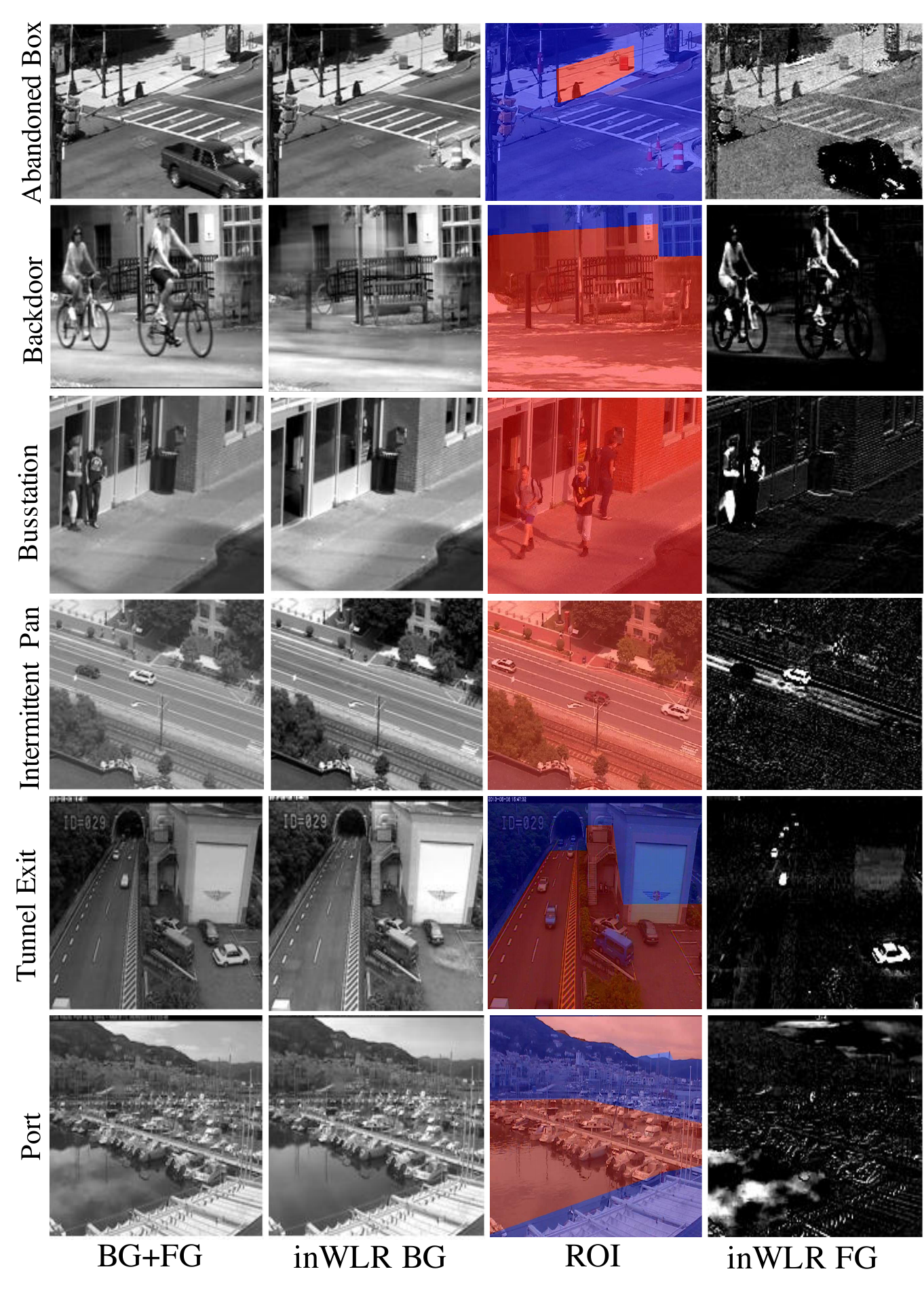}
	\caption{\small{Qualitative results of inWLR on selected sequences of CDnet 2014 dataset.}}\label{CDnet_result1}
\end{figure}
\subsubsection{Comparion with RPCA}
Now we compare Algorithm \ref{bg_wlr} and Algorithm \ref{inwlr} with the RPCA. For this purpose we consider APG and iEALM. For both algorithms we set $\lambda=1/\max\{m,n\}$, and for iEALM we choose $\mu=1.25/\|A\|_2$ and $\rho=1.5$, where $\|A\|_2$ is the spectral norm (maximum singular value) of $A$. From Figure \ref{rpca} it is clear that when the foreground is static for a few frames both APG and iEALM can not remove the static foreground object. Moreover, to process 600 frames of resolution $144\times 176$, iEALM and APG took 501.46 seconds and 572.49 seconds, respectively. 

\subsubsection{Comparison with other state-of-the-art-methods}
Next we compare Algorithm \ref{bg_wlr} and Algorithm \ref{inwlr} with other state-of-the-art background estimation methods, such as, GRASTA (with different subspample ratio), incPCP, ReProCS, PRMF, GFL, and DECOLOR. We do not include GOSUS \cite{gosus} as it is very similar to GRASTA. For GRASTA, we set the subsample percentage $s$ at 0\%, 10\%, 20\%, and at 30\% respectively,
estimated rank 60, and keep the other parameters the same as those in \cite{grasta}. We use 200 background frames of the Basic sequence for initialization of ReProCS. According
to \cite{incpcp}, the initialization step can be performed incrementally.
For the Stuttgart sequence, the algorithm uses
the first video frame for initialization. PRMF and DECOLOR are unsupervised algorithms. We set the target rank for PRMF to 5 and the other parameters are kept same as in the software package. For DECOLOR we use the static camera interface of the code and the parameters are kept same as they are mentioned in the software package. The ROC curves in Figures \ref{roc_wlr} and \ref{roc_inwlr} to demonstrate that our proposed algorithms outperform other methods. In Figure \ref{msim_wlr} we present the mean SSIM of the recovered foreground frames and Algorithm \ref{inwlr} outperforms ReProCS. In contrast, incPCP has similar mean SSIM. Moreover, in Figure \ref{420st} all methods except GRASTA appear to perform equally well on the {\tt Basic} scenario. However, when the foreground
is static (as in frames 551-600 of the {\tt Basic} scenario) neither incPCP, PRMF, nor, DECOLOR can capture the static foreground object, thus result the presence of the static car as a part of the background (see Figure \ref{600st}). For supervised GFL model, we use 200 frames from each scenario ({\tt Basic} and {\tt Waving Tree}) for training purpose. The background recovered and the SSIM map in Figures \ref{600st} show that GFL provides a comparable reconstruction and can effectively remove the static car (also see Figure \ref{WT_32}). However, it is worth mentioning that inWLR is extraordinarily time efficient compare with the GFL model. We also note that ReProCS is a robust method in removing several background challenges. Although, ReProCS foreground has more false positives than inWLR. This explains why ReProCS suffers in quantitative evaluation (see Figure \ref{123st} and \ref{WT_32}).
\begin{figure}
	\centering  
	\includegraphics[width=\linewidth, height=2.9in]{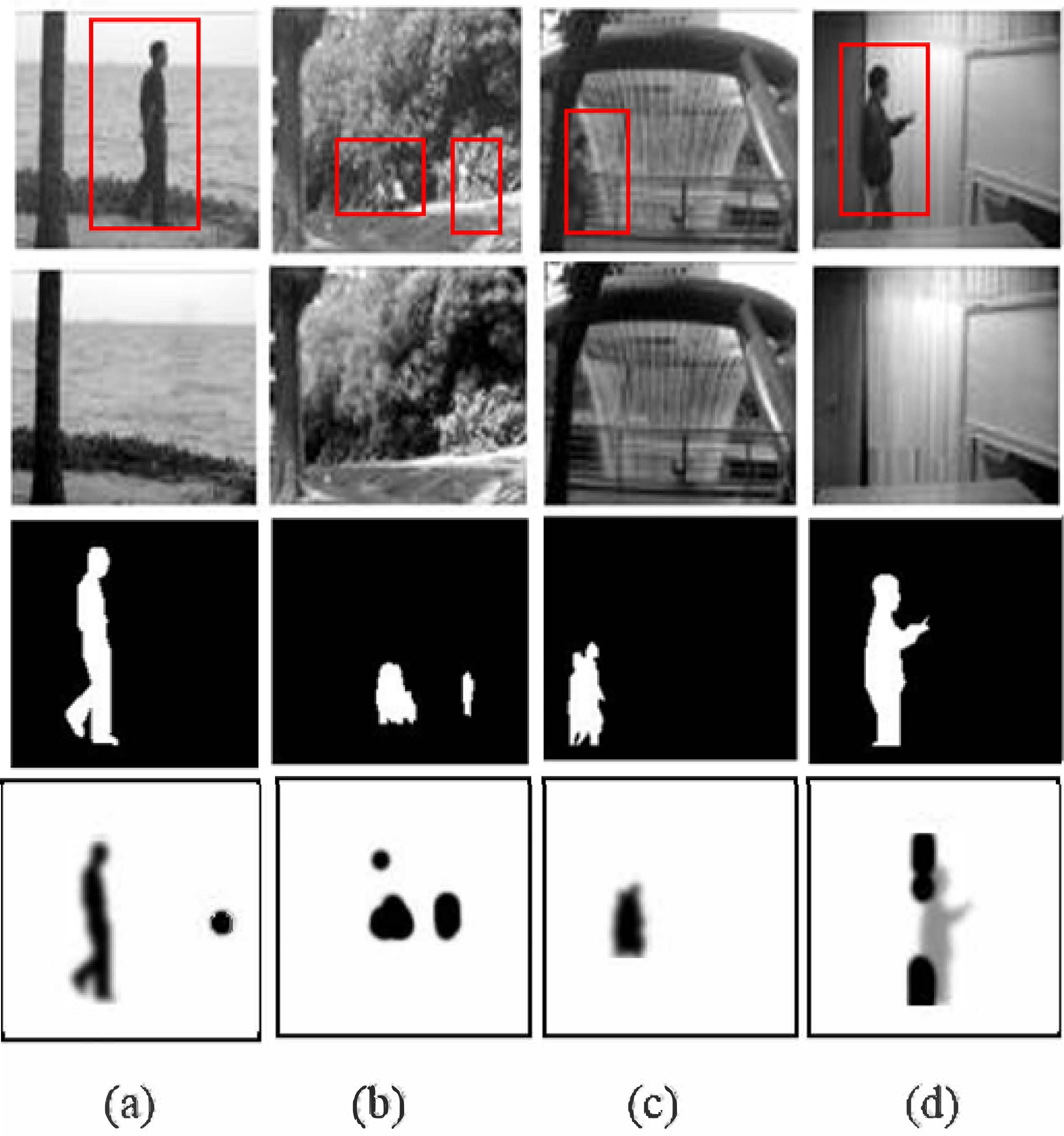}
	\caption{\small{SSIM index map of:~(a)~{\it Water Surface},~(b)~{\it Waving tree},~(c)~{\it Fountain}, and~(d)~{\it Meeting Room}.~Top to bottom: Original, background estimated by Algorithm \ref{bg_wlr},~ground truth frame~(size $64\times 80$), SSIM index map~(size $54\times 70$)~of Algorithm \ref{bg_wlr}.~The MSSIM are 0.9851, 0.9082, 0.9940,~and~0.9343, respectively.}}\label{dyn_for}
\end{figure}
 \begin{table}
\begin{center}
\begin{tabular}{|l|c|c|c|}
\hline
Video & MSSIM & MSSSIM & PSNR~(dB) \\
\hline
{\tt Abandoned Box}& 0.9836 & 0.9980 &58.6752 \\
\hline
{\tt Backdoor} &0.9836 & 0.9978 & 60.4562  \\
\hline
{\tt Busstation} &0.9892 & 0.9987 & 57.7467\\
\hline
{\tt Intermittent Pan} &0.9836 &0.9978 & 57.7169  \\
\hline
{\tt Tunnel Exit} & 0.9832& 0.9977 & 57.6781\\
\hline
{\tt Port}&0.9833 &0.9978 &57.6868\\
\hline
{\tt Fountain} &0.9886 &0.9985 &57.6812\\
\hline
{\tt Overpass} &0.9847 &0.998 & 57.676\\
\hline
{\tt Fountain 2} &0.9934 &0.9992 &57.6777\\
\hline
{\tt Canoe} &0.9858 & 0.9981 &57.678\\
\hline
{\tt Fall} &0.9853 &0.9980 &57.7461\\
\hline
\end{tabular}
\end{center}
\caption{\small{Quantitative performance of inWLR on CDNet 2014 dataset~\cite{sbi_a}}.}\label{cdnet_data}
\end{table}
\vspace{-0.1in}

\subsection{Performance on SBI dataset}

SBI dataset comprises 14 image sequences and they come with 14 background ground truths. Therefore, we can validate the background recovered by our inWLR algorithm against the background ground truth. For this purpose, we use all four quantitative metrics: area under the ROC curve, SSIM, PSNR, and MSSSIM. To calculate MSSSIM for the SBI dataset, we use a Gaussian window of size $9\times9$ and standard deviation ($\sigma$) 1.5. In Figure \ref{SBI_result} we present the qualitative performance of inWLR on 7 different sequences of the SBI dataset and we refer to Table \ref{sbi_data} for quantitative measures. The missing values under the ``Area under ROC curve" column of Table \ref{sbi_data} is due to {\tt NaN} values corresponding the false positive rate (FPR) of the predictive analysis. We note that the backgrounds recovered by inWLR sometimes have a minor ghosting effect (see red bounding boxes in Figure \ref{SBI_result}). However, the average PSNR and MSSSIM of inWLR on 7 sequences of the SBI dataset are 74.3221 and 0.9995, respectively. In contrast, the spatially coherent self-organizing background subtraction (SC-SOBS1) algorithm has the highest average PSNR (35.2723) and MSSSIM (0.9765) on the entire SBI dataset \cite{sobs,SBI_web}. Recently, the supervised online algorithms, such as, IRLS and Homotopy proposed by Dutta and Richt\'{a}rik \cite{duttarichtarik}, have average MSSSIM on the SBI dataset as 0.9975 and 0.9987, respectively. 

\begin{figure}
    \centering
    \includegraphics[width=0.5\textwidth,height=0.7in]{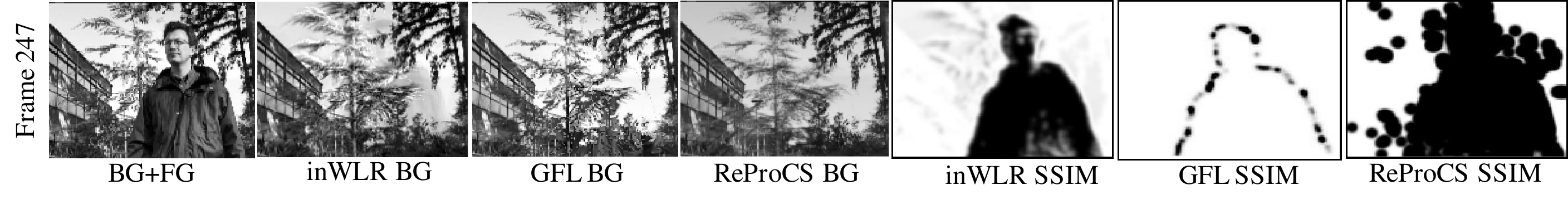}
    \caption{\small{Waving Tree,frame size [120,160]. ReProCS and GFL
use 220 and 200 pure background frames respectively as
training data. The MSSIM for inWLR, GFL, and ReProCS are 0.9592, 0.9996, and
0.5221, respectively. inWLR and GFL recover superior quality background.}}
    \label{WT_32}
\end{figure}
\begin{figure}
    \centering
    \includegraphics[width=3.1in, height= 1.3in]{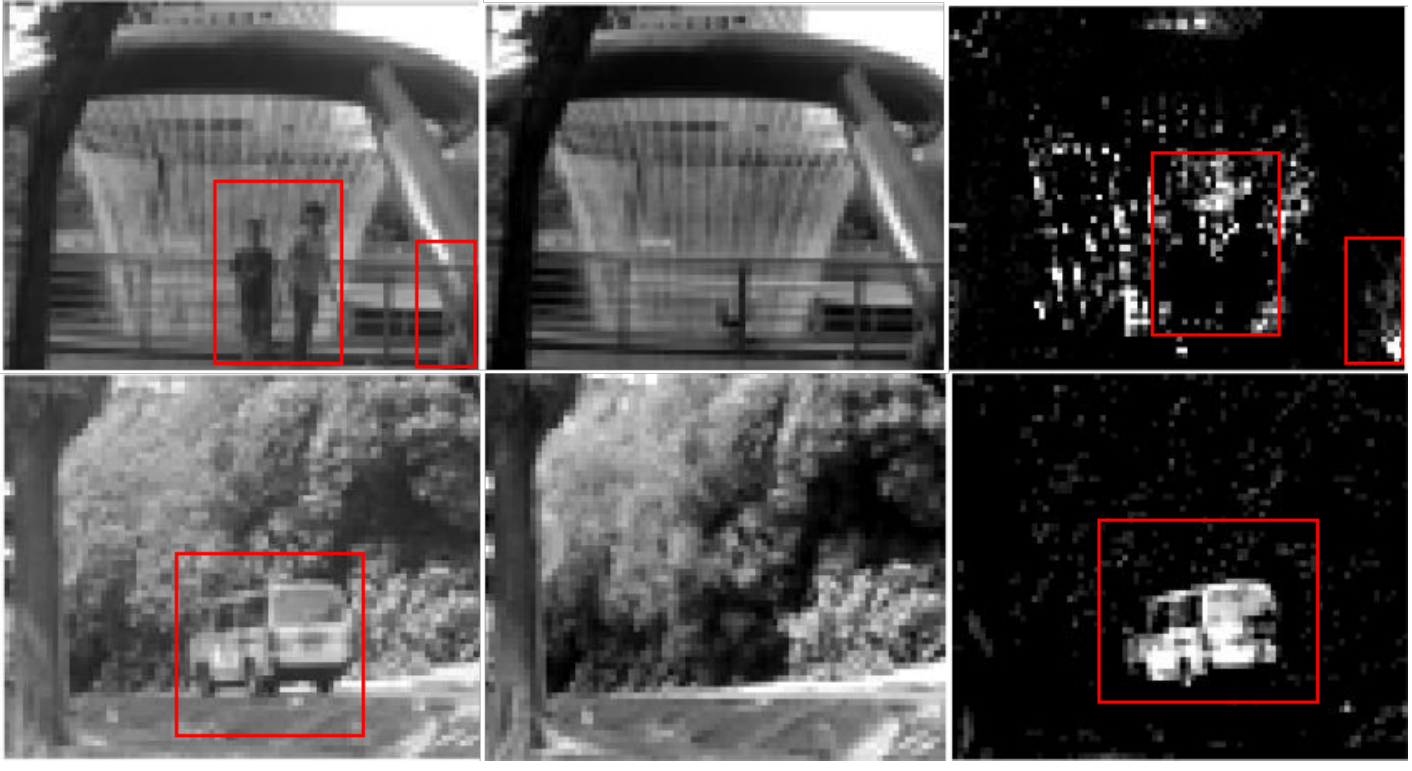}
    \caption{Top to bottom:~{\tt Fountain} with $p=5$, {\tt Campus} with $p=6$.~Left to right:~Original, inWLR background,~and~inWLR foreground.}
    \label{inwlr_bg}
\end{figure}
\vspace{-0.1in}

\subsection{Performance on CDNet2014 dataset}
The CDNet 2014 dataset contains a total 11 different video categories. Additionally, each video comes with region of interest (ROI) image which defines the region(s) in a video with foreground movement. The ROI is the red colored part of the frame (with static camera movement). We choose a total 11 sequences from the CDNet 2014 dataset and test the performance of inWLR algorithm both qualitatively and quantitatively. We use SSIM, PSNR, and MSSSIM for quantitative measure. We refer to Figure \ref{CDnet_result1} for the first set of qualitative results. The average PSNR of inWLR on 11 video sequences of CDNet 2014 dataset is 58.0381 dB.  

\vspace{-0.1in}

\subsection{Dynamic background: a case study}
\begin{figure}
	\centering  
	\includegraphics[width=\linewidth, height = 3.5in]{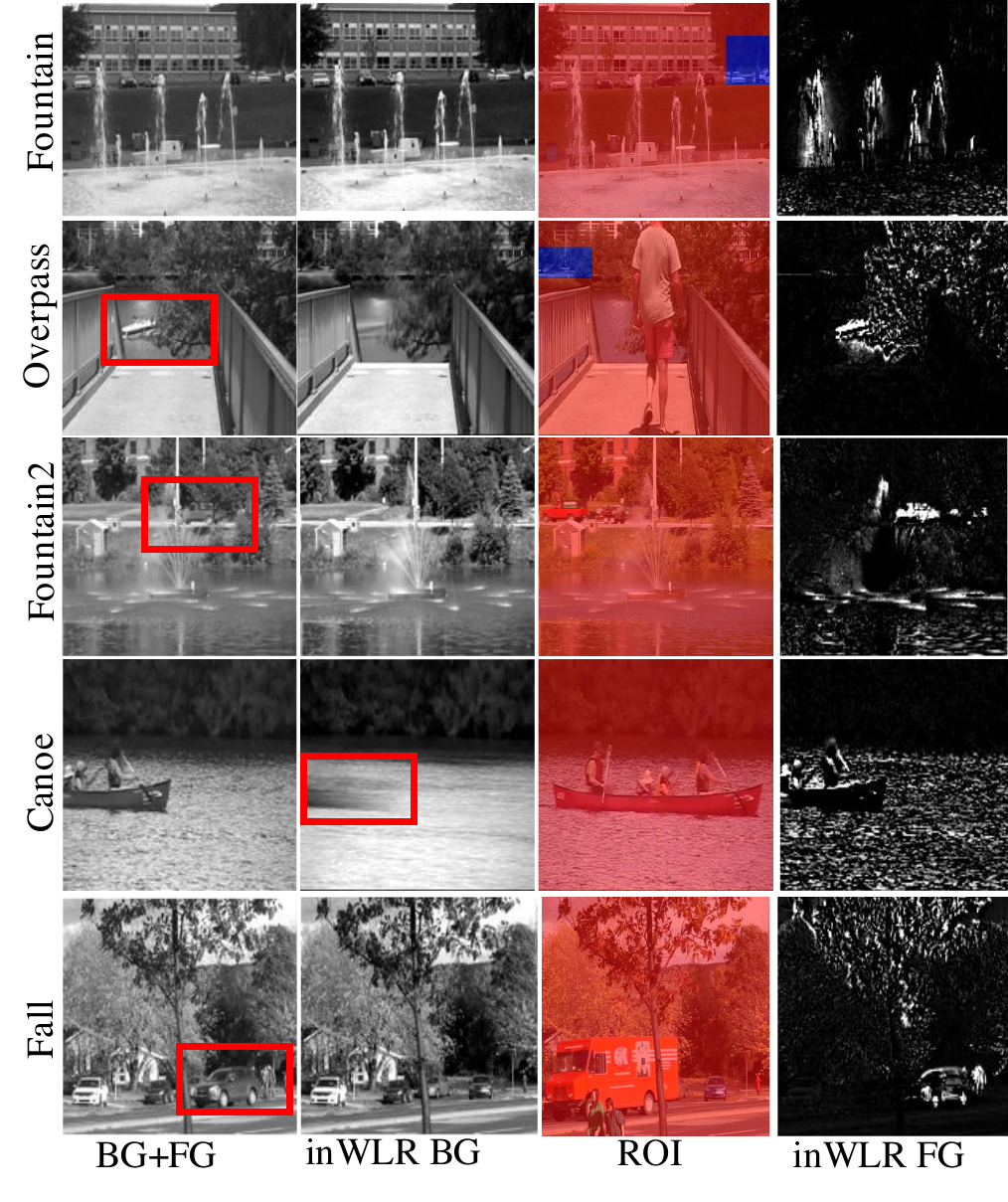}
	\caption{\small{Qualitative results of inWLR on selected sequences of CDnet 2014 dataset with dynamic background. Only in the {\tt Canoe} sequence, inWLR has some ghosting effect of the foreground object in the background.}}\label{CDnet_result2}
\end{figure}

%Describe the problem and propose our solution. Compare with other methods. Compare with other dynamic detection methods.
Dynamic background objects are a potential challenge for background modeling algorithms as it is natural for an algorithm to consider them as a part of the foreground. 
To demonstrate the power of our method on more complex
data sets containing dynamic foreground, we perform
extensive qualitative and quantitative analysis on the Li data
set \cite{lidata}, Wallflower data set \cite{wallflower}, and CDNet 2014 dataset \cite{cdnet}. Both our algorithms are capable of detecting the dynamic background objects, although Algorithm \ref{inwlr} is a more natural choice as it deals with the video sequence in an incremental manner. First, in Figure \ref{dyn_for}, we show the performance of Algorithm \ref{bg_wlr}. The SSIM index map on all four
recovered foreground indicates that WLR performs consistently
well on the video sequences containing dynamic
background. Next, in Figure \ref{WT_32}, we compare inWLR against GFL and ReProCS
on 60 frames of {\tt Waving Tree} sequence. In Figure \ref{inwlr_bg} we show the performance of inWLR on two data sets with dynamic background and semi-static foreground. Finally, in Figure \ref{CDnet_result2} we present the performance of inWLR on the CDNet 2014 dataset with dynamic background. We also provide the quantitative results in Table \ref{cdnet_data}. 
\vspace{-0.1in}

\section{Conclusion}
In this paper we proposed two novel algorithms for background
estimation. We demonstrated how
a properly weighted Frobenius norm can be made robust to
the outliers, similarly to the $\ell_1$ norm in other state-of-the-art background estimation algorithms. Both of our algorithms adaptively determine the background
frames without requiring any prior estimate. Furthermore,
our batch-incremental algorithm does not require much storage and allows slow
changes in the background. Our extensive qualitative and
quantitative comparison on real and synthetic video sequences
demonstrate the robustness of our algorithms. The batch
sizes and the parameters in our incremental algorithm are still empirically
selected. Therefore, in future we plan to propose a more robust estimate
of the parameters. %and explore the possibilities that our
%algorithm can handle videos of more dynamic background. 
Like all other algorithms in the literature, we have some limitations as well. Although we are not explicitly required to use pure training background frames, it is mandatory that the video has availability of some pure background frames. Then our algorithm can automatically detect them and use them efficiently in background modeling. Otherwise it will detect the frames with least foreground movements and approximately preserve them as a contender of the background. In the later case the constructed background can be defective.

\vspace{-0.1in}

\section*{Acknowledgment}
This work was supported by the King Abdullah University of Science and Technology (KAUST) Office of Sponsored Research.

\ifCLASSOPTIONcaptionsoff
  \newpage
\fi



%\bibliographystyle{IEEEtran}
%\clearpage
\bibliographystyle{IEEEtran}
\bibliography{egbib_updated}

%\begin{thebibliography}{1}
%\end{thebibliography}


\begin{IEEEbiography}[{\includegraphics[width=1in,height=1.25in,clip,keepaspectratio]{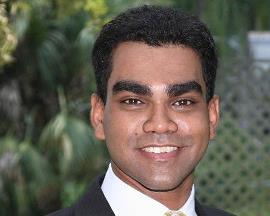}}]{Aritra Dutta} received his B.S. in mathematics from the Presidency College, Calcutta, India, in 2006. He received his M.S. in mathematics and computing from the Indian Institute of Technology, Dhanbad, in 2008 and a second M.S. in mathematics from the University of Central Florida, Orlando, in 2011. He received his PhD in mathematics from the University of Central Florida in 2016. Dr. Dutta is currently working as a Postdoctoral Fellow at the Visual Computing Center at the Division of Computer, Electrical and Management Sciences \& Engineering, King Abdullah University of Science and Technology (KAUST), Saudi Arabia. His research interests include weighted and structured low-rank approximation of matrices, convex, nonlinear, and stochastic optimization, numerical analysis, linear algebra, and deep learning. In addition, he works on the applications of image and video analysis in computer vision and machine learning.
\end{IEEEbiography}

\begin{IEEEbiography}[{\includegraphics[width=1in,height=1in,clip,keepaspectratio]{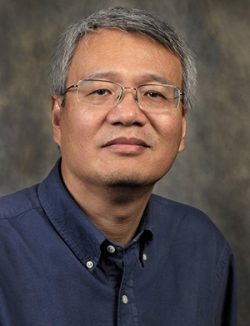}}]{Xin Li}
Dr. Xin Li is Professor and Chair of Mathematics at the University of Central Florida. He received his B.S. and M.S. degrees from Zhejiang University in 1983 and 1986, respectively, and his PhD in Mathematics from the University of South Florida in 1989. Dr. Li's research interests include Approximation Theory and its applications in scientific computing, machine learning, and computer vision.
\end{IEEEbiography}
%\newpage
\begin{IEEEbiography}[{\includegraphics[width=1in,height=1.25in,clip,keepaspectratio]{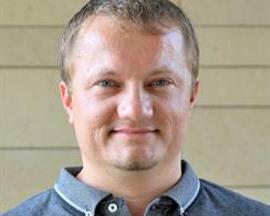}}]{Peter Richt\'{a}rik} is an Associate Professor of Computer Science and Mathematics at KAUST and an Associate Professor of Mathematics at the University of Edinburgh. He is an EPSRC Fellow in Mathematical Sciences, Fellow of the Alan Turing Institute, and is affiliated with the Visual Computing Center and the Extreme Computing Research Center at KAUST. Dr. Richtarik received his PhD from Cornell University in 2007, and then worked as a Postdoctoral Fellow in Louvain, Belgium, before joining Edinburgh in 2009, and KAUST in 2017. Dr. Richtarik's research interests lie at the intersection of mathematics, computer science, machine learning, optimization, numerical linear algebra, high performance computing, and applied probability. Through his recent work on randomized decomposition algorithms (such as randomized coordinate descent methods, stochastic gradient descent methods and their numerous extensions, improvements and variants), he has contributed to the foundations of the emerging field of big data optimization, randomized numerical linear algebra, and stochastic methods for empirical risk minimization. Several of his papers attracted international awards, including the SIAM SIGEST Best Paper Award and the IMA Leslie Fox Prize (2nd prize, three times). He is the founder and organizer of the Optimization and Big Data workshop series.\end{IEEEbiography}


%\vfill


\end{document}